\documentclass[10pt,twocolumn,letterpaper]{article}

\usepackage[pagenumbers]{cvpr}      %

\usepackage{cuted} %

\usepackage{currfile} %

\usepackage{caption} %

\newcommand{\red}[1]{{\color{red}#1}}

\usepackage{booktabs}
\usepackage{multirow}
\usepackage{makecell}
\usepackage{xspace}
\usepackage{amsmath,amssymb}

\newcommand{\bench}{MedVidBench-v2\xspace}

\newcommand{\best}[1]{\textbf{#1}}
\newcommand{\method}{LQR\xspace}  %

\usepackage{colortbl}  %
\newcommand{\gain}{\cellcolor{green!18}}
\newcommand{\loss}{\cellcolor{red!15}}

\definecolor{cvprblue}{rgb}{0.21,0.49,0.74}
\usepackage[pagebackref,breaklinks,colorlinks,allcolors=cvprblue]{hyperref}

\def\confName{CVPR}
\def\confYear{2027}

\title{Evidence-Aligned Benchmarking and Learning\\for Medical Video Language Models}

\author{Zhongpai Gao$^{*}$\hspace{0.5em} Anwesa Choudhuri\hspace{0.5em} Meng Zheng\hspace{0.5em} Benjamin Planche\hspace{0.5em} Terrence Chen\hspace{0.5em} Ziyan Wu\\
United Imaging Intelligence, Boston, MA, USA
}

\begin{document}
\twocolumn[{%
\renewcommand\twocolumn[1][]{#1}%
\maketitle
\begin{center}
\vspace{-1.5em}
\includegraphics[width=\linewidth]{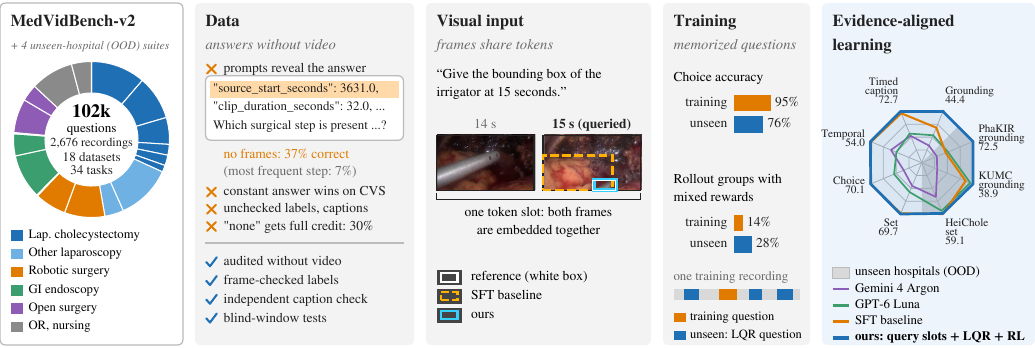}
\captionof{figure}{\textbf{Evidence-aligned benchmarking and learning.} Left: \bench. Middle: benchmarks reward answers given without video, frames share tokens, and training questions are memorized; we respond with an audit, query-slot inputs, and label-matched question resampling (\method). Right: comparison with the SFT baseline and commercial models; each axis ends at its best score.}
\label{fig:teaser}
\end{center}%
}]
\begingroup\renewcommand\thefootnote{*}\footnotetext{Corresponding author.}\endgroup
\begin{abstract}
A medical video language model should answer from the visual evidence its question asks about. We find three places where this breaks down and address each with an approach we call MedEvidU. First, benchmarks reward answers that ignore the video: on one dataset, the position of a clip within its recording predicts the surgical step in 37\% of questions, and a constant answer beats every trained model on a standard safety-assessment metric. We build \bench, 34 tasks from 18 surgical and clinical video datasets, in which prompts and metrics are audited with answers given without video, and which tests unseen hospitals and windows cut blindly from whole recordings. Second, common video encoders merge consecutive frames into shared tokens, so a question about one frame is answered from a blend of two. Query-slot inputs give each queried frame its own tokens at almost no extra cost. Third, a fine-tuned model has memorized its training questions, which leaves little to learn from. Label-matched question resampling (\method) adds unseen questions about the same recordings while keeping the answer distribution fixed, in a final supervised epoch and in reinforcement learning. Together, these raise the grounding IoU of a 4B model from 30.4 to 44.4, above a supervised 27B model, and instrument grounding at an unseen hospital from 33.9 to 72.5. On blindly cut windows, event-presence accuracy improves by 5.5 points and localization F1 by 8.1 points.
Project page: \url{https://gaozhongpai.github.io/MedEvidU/}
\end{abstract}

\section{Introduction}
\label{sec:intro}

A medical video language model should answer from the visual evidence its question asks about: the interval of an action, the location of a tool, the current surgical phase, or whether a safety milestone such as the Critical View of Safety (CVS) has been reached. Recent models handle all of these through one question-answering interface, trained on many surgical and clinical datasets with multi-task reinforcement learning~\cite{su2026medgrpo,huang2026surgmllm}. A high score, however, does not show that the model used the evidence. We find three places where the two come apart: the data, the visual input, and the training questions (\cref{fig:teaser}). Addressing each is what we call \emph{evidence-aligned} benchmarking and learning (MedEvidU).

The first is the data. A benchmark can reward answers that never use the video, through its prompts and through its metrics. When we ask models to answer without any frames, the position of a clip within its recording predicts the surgical step in 37\% of MultiBypass140 questions, against 7\% for the most frequent step. Some prompts list only a few candidate answers, which rules out most wrong ones. Metrics leak as well: a constant ``never met'' answer beats every trained model on the standard CVS trajectory metric, and 30\% of temporal questions give full credit to the answer \texttt{none}. Curated clips add a further cue, because a localization clip is cut around its event. Data of this kind rewards shortcuts over video understanding~\cite{geirhos2020shortcut,buch2022revisiting}.

We introduce \bench, with 34 tasks from 18 surgical and clinical video datasets, built so that a score requires the video. We audit prompts and metrics with answers given without frames: leaking context is removed, every recognition question lists the full label vocabulary, trajectories are scored with class-balanced accuracy, and event presence is reported separately from localization. A blind-window test cuts whole recordings into windows without looking at where events lie, as a deployed system would, so the model must decide whether an event occurs before saying where. Four suites from unseen hospitals measure transfer. Labels and caption references are checked against the frames, and all splits are by recording.

The second is the visual input. Clean data does not guarantee that the model sees the right frame. Many video encoders embed two consecutive frames into one set of tokens~\cite{arnab2021vivit,tong2022videomae}, including the Qwen2.5-VL, Qwen3-VL, and GLM-4.1V/4.5V families~\cite{bai2025qwen25vl,bai2025qwen3vl,glmv2025glm41v}. A question that asks for a box at one time is then answered from a blend of that frame and its neighbor. \emph{Query-slot inputs} repeat the queried frame in place of its neighbor, so its tokens describe that frame alone. The change needs only the frames named in the prompt, adds almost no tokens, and applies in both training and inference.

The third is training. After supervised fine-tuning (SFT), a model has largely memorized its training questions. Repeating them teaches little, and in reinforcement learning most sampled answers to a memorized question earn the same reward, which gives no gradient. \emph{Label-matched question resampling} (\method) supplies unseen questions and newly cut clips from the same training recordings, without new annotation. Because unseen questions have different answer statistics, \method keeps the number of questions per task and the answer distribution of the original training set. We use it for a final supervised epoch and for reinforcement learning whose rewards come from the audited benchmark metrics.

Our contributions are:
\begin{itemize}
\item \textbf{Benchmark.} \bench, a medical video benchmark whose prompts and metrics are audited against answers that ignore the video, with blind-window and unseen-hospital tests and frame-checked references.
\item \textbf{Method.} \emph{Evidence-aligned learning}, which combines query-slot inputs with \method in a final supervised epoch and in reinforcement learning.
\item \textbf{Results.} Against a baseline with equal supervised training, a 4B model improves grounding IoU from 30.4 to 44.4, above a supervised 27B model, and instrument grounding at an unseen hospital from 33.9 to 72.5.
\end{itemize}

\section{Related Work}
\label{sec:related}

\noindent\textbf{Medical video benchmarks.}
Public surgical datasets cover phase and step recognition~\cite{twinanda2017endonet,twinanda2016m2cai,lavanchy2024multibypass,wang2022autolaparo}, action triplets~\cite{nwoye2022rendezvous}, tool tracking and segmentation~\cite{nwoye2025cholectrack20,alabi2025cholecinstanceseg,psychogyios2023sarrarp50}, safety assessment~\cite{rios2023cholec80cvs,murali2025endoscapes}, skill assessment~\cite{gao2014jigsaws}, and multi-center workflows~\cite{wagner2023heichole}, each with its own labels and metric. Video language benchmarks cast such tasks as question answering: answer localization in instructional videos~\cite{gupta2022medvidqa}, questions on surgical lecture videos~\cite{li2024llavasurg,wang2025surgvidlm}, surgeon-verified reasoning~\cite{perez2026sureon}, and the multi-task MedVidBench~\cite{su2026medgrpo} (\cref{tab:lineage}). \bench evaluates one model on all of these task types under a shared protocol, and adds what these benchmarks do not test: whether a score can be earned without the video, and whether it holds on blindly cut windows and at unseen hospitals.

\noindent\textbf{Shortcuts and benchmark validity.}
Models exploit regularities that bypass the intended input~\cite{geirhos2020shortcut}, and single-frame and language-only baselines expose this in video understanding~\cite{buch2022revisiting}. We apply the same test to medical video and extend it from the questions to the metrics: the audit finds prompts that reveal the answer and metrics that credit a constant answer, and \bench removes both.

\noindent\textbf{Frame-level evidence in video models.}
Video transformers often embed short groups of consecutive frames~\cite{arnab2021vivit,tong2022videomae}, and widely used video language models keep two-frame patches~\cite{bai2025qwen25vl,bai2025qwen3vl,glmv2025glm41v,qwen2026qwen35}. A question about one frame is then answered from tokens shared with its neighbor. Query-slot inputs remove this mismatch without changing the model; encoders that embed frames separately do not need them.

\noindent\textbf{Medical video language models and reinforcement learning.}
MedGRPO~\cite{su2026medgrpo} trains one model on many medical video tasks with task-specific rewards, SurgMLLM~\cite{huang2026surgmllm} links workflow reasoning to pixel-level grounding, and report generation has been studied for cataract surgery~\cite{yu2026lensnarrate}. Group-relative policy optimization (GRPO)~\cite{shao2024deepseekmath} and its DAPO variant~\cite{yu2025dapo} optimize verifiable rewards without a learned critic, and related methods extend this to video reasoning~\cite{feng2025videor1}. When reinforcement learning reuses the supervised training questions, most sampled answers to a question earn the same reward, and DAPO discards such questions. \method is complementary: it supplies unseen questions on which answers differ, and also helps supervised training (\cref{sec:rl}).

\section{The \bench Benchmark}
\label{sec:benchmark}

\begin{table}[t]
\centering
\caption{\textbf{Medical video QA benchmarks.} QA: train / test. Blind windows reuse in-domain test videos (\cref{tab:data_composition}). Ans.: L label, S set, T span, B box, C text. Ver.: verified test.}
\label{tab:lineage}
\scriptsize
\setlength{\tabcolsep}{2.2pt}
\begin{tabular}{@{}l r r c c c c@{}}
\toprule
Benchmark & Videos & Tasks & QA & Ans. & Ver. & OOD \\
\midrule
MedVidQA~\cite{gupta2022medvidqa} & 899 & 1 & 2,710 / 155 & T & \checkmark & -- \\
Surg-QA~\cite{li2024llavasurg} & 2,151 & 11 & 90K / 4,359 & C & -- & -- \\
SVU-31K~\cite{wang2025surgvidlm} & 1,882 & -- & 31K total & C & -- & -- \\
SUREON~\cite{perez2026sureon} & 2,356 & 10 & 76,061 / 340 & L, C & \checkmark & -- \\
MedVidBench~\cite{su2026medgrpo} & 626 & 8 & 45,260 / 6,245 & L, T, B, C & -- & -- \\
\midrule
\bench (ours) & 2,676 & 34 & 77,261 / 16,361 & L, S, T, B, C & \checkmark & \checkmark \\
\quad + blind windows & (90) & 9 & -- / 8,292 & L, S, T, C & & \\
\quad + OOD & 81 & 11 & -- / 4,260 & L, S, T, B, C & & \\
\bottomrule
\end{tabular}
\end{table}

\bench poses surgical and clinical video understanding as question answering over clips. Each question pairs a clip with a task specification that states the answer format and, for recognition tasks, the full list of candidate labels. Compared with prior medical video question-answering benchmarks (\cref{tab:lineage}), \bench covers the widest range of tasks and answer types. Its design follows one requirement: a score must be earned from the video. We audit prompts and metrics with answers given without video (\cref{sec:shortcuts}), test on blindly cut windows and at unseen hospitals (\cref{sec:realistic}), and check every reference against the frames (\cref{sec:captions}).

\subsection{Sources and tasks}
The benchmark draws on 18 public datasets: laparoscopic cholecystectomy~\cite{twinanda2017endonet,nwoye2022rendezvous,rios2023cholec80cvs,murali2025endoscapes,nwoye2025cholectrack20,alabi2025cholecinstanceseg,twinanda2016m2cai}, gastric bypass~\cite{lavanchy2024multibypass}, hysterectomy~\cite{wang2022autolaparo}, robot-assisted prostatectomy~\cite{psychogyios2023sarrarp50,prostatd2025}, endoscopic submucosal dissection~\cite{wang2024copesd}, colonoscopy~\cite{biffi2024realcolon,ma2021ldpolypvideo,surgvivqa2025}, open and egocentric surgery~\cite{goodman2021avos,fujii2024egosurgery}, operating-room activity~\cite{ozsoy2025egoexor}, robotic skill~\cite{gao2014jigsaws}, and nursing procedures~\cite{hu2023nurvid}. Their annotations support 34 tasks, which we group into families by answer type and metric (\cref{tab:families}).

\begin{table}[t]
\centering
\caption{\textbf{Task families of \bench.} Scores average over datasets within a task and over tasks within a family; $n$: test questions in the headline scope.}
\label{tab:families}
\scriptsize
\setlength{\tabcolsep}{3pt}
\begin{tabular}{@{}l r p{3.15cm} l@{}}
\toprule
Family & $n$ & Tasks & Metric \\
\midrule
Temporal & 3,222 & action / instrument / gesture localization, visibility intervals & F1 @ tIoU 0.5 \\
Spatial & 1,584 & spatio-temporal grounding & box mIoU \\
Choice & 4,404 & phase, step, action recognition; next action / phase / step; histology; lesion; colon scene questions & accuracy \\
Set & 3,093 & action, instrument, segment, triplet sets; scene graphs; adverse events & set F1 \\
Ordered & 300 & step sequences & 1 $-$ norm.\ edit dist. \\
Component & 587 & CVS criteria (frame, clip); skill & class-bal.\ accuracy \\
Trajectory & 137 & CVS criterion onset over a video & class-bal.\ accuracy \\
Timed caption & 1,436 & dense and multi-view dense captions & F1 @ tIoU 0.5; judge \\
Caption & 338 & summaries, region captions & rubric judge \\
\bottomrule
\end{tabular}
\end{table}

\bench has 77,261 training, 8,454 validation, and 16,361 test questions over 2,676 recordings, balanced across tasks and datasets with a limit on the share of any single recording (\cref{sec:supp_data}, \cref{fig:composition}). Headline results use the 15,101 test questions from task--dataset pairs that have at least 60 questions from at least two recordings. Every dataset is split by recording. These splits differ from some official source splits, so results are not comparable with numbers reported on those.

\subsection{Auditing without video}
\label{sec:shortcuts}
We audit the benchmark with answers that do not use the video: answers of models that receive no frames, the literal answer \texttt{none}, and the most frequent training answer for each task and dataset (\cref{tab:main}, \cref{fig:videofree}). A prompt or metric fails when such an answer scores near a trained model.

\noindent\textbf{Prompts that reveal the answer.}
The position of a clip within its recording predicts workflow labels: on MultiBypass140, position alone gives the correct step in 37\% of questions, against 7\% for the most frequent step. \bench prompts therefore carry no recording position or recording-clock time, only the clip-relative times that temporal tasks need. Some source prompts also list a shortened set of candidates that excludes valid alternatives, or name the current phase in a question about its steps. In \bench, every recognition question lists the full task vocabulary and carries no such context.

\noindent\textbf{Answer distributions.}
A question should not reveal its answer through its format or source. All temporal tasks therefore use one format for events that occur and events that do not, and 22--34\% of temporal training answers in every task and dataset are empty. Set-valued tasks are sampled so that no constant answer scores well (\cref{sec:supp_data}).

\noindent\textbf{Metrics.}
\cref{tab:families} lists the metric of each family. Two metrics fail the audit. The standard CVS trajectory metric averages accuracy over the onset and final status of each criterion. Most criteria are never met within a test video, so a constant ``never met'' answer scores 66.4, above every trained model. We replace it with the mean accuracy on met and never-met references, under which any constant answer scores 50. Temporal F1 matches predicted and reference intervals one-to-one at temporal IoU of at least 0.5, and an empty prediction for an empty reference scores 1. Since 30\% of temporal test questions have empty references, \texttt{none} scores 35.9 on the curated test and 70.6 on blind windows (\cref{sec:realistic}), where trained models score 78--79. Temporal F1 thus mixes two abilities, deciding whether an event occurs and finding where, and we report them separately. \emph{Presence} is the mean of event recall (an interval is returned when an event occurs) and empty accuracy (\texttt{none} is returned otherwise); any constant decision scores 50. \emph{Localization F1} is the temporal F1 on questions whose reference contains an event. Timed-caption F1 also rewards clip structure, since equal-length segments score 50.2, so we read it together with caption quality and the blind-window test. In the other families, trained models are clearly separated from answers that ignore the video.

\subsection{Test suites}
\label{sec:realistic}

\noindent\textbf{Curated test.}
The main test set contains clips built around annotations, in the same way as the training clips but from held-out recordings.

\noindent\textbf{Unseen hospitals.}
Four out-of-distribution (OOD) suites come from hospitals absent from training.
\emph{HeiChole}~\cite{wagner2023heichole} has 1,807 questions from 24 cholecystectomies at two hospitals, covering action and instrument sets, phase recognition, next-phase anticipation, and dense captions.
\emph{MultiBypass140 Strasbourg}~\cite{lavanchy2024multibypass} has 1,515 questions from 14 gastric-bypass recordings; training uses only the Bern recordings of that dataset.
\emph{KUMC}~\cite{li2021kumc} has 48 polyp-grounding questions from a third colonoscopy center.
\emph{PhaKIR}~\cite{rueckert2025phakir} has 890 questions from eight complete cholecystectomies at three further centers, covering phases, anticipation, instrument sets and grounding, and dense captions.

\noindent\textbf{Blind windows.}
Curated clips reveal where events are: a localization clip encloses its event, and a dense-caption clip is centered on a phase change. Splitting a hysterectomy clip into two equal halves scores 77.4 on dense-caption timing, against 78.3 for a trained model. The blind-window test therefore tiles each test recording with fixed-length windows from a random offset, without consulting the annotations, and regenerates the answers from the native timelines. It contains 8,292 questions and keeps the window lengths and frame sampling of the curated test. Localization questions ask about every label of the recording in every window, so most windows contain no event, and the model must first decide whether the event is present. Training includes 1,622 questions cut the same way from training recordings (\cref{sec:supp_realistic}).

\subsection{Checked references}
\label{sec:captions}

\noindent\textbf{Captions.}
Most source captions were written by language models and never compared with the video. \bench replaces them: 18,786 training, 1,040 validation, and 1,428 test captions are rewritten from the frames, and an independent pass checks each one, with the target box shown for region captions. Each test reference is drafted twice and reconciled against the frames, and a different model family audits a sample (\cref{sec:supp_data}).

\noindent\textbf{Labels.}
Structured answers are checked against the native annotations and the frames they describe. Endoscapes provides CVS labels for every frame, but experts rated only 19\% of them; the rest are copied from neighboring keyframes. We ask frame-level CVS questions only at rated keyframes and use the majority label of the raters, who disagree on at least one criterion in 63\% of test frames. For phase and step recognition, every evaluation answer equals the native label at the last frame shown, and each task uses one fixed vocabulary.

\section{Evidence-Aligned Learning}
\label{sec:method}

\begin{figure*}[t]
\centering
\includegraphics[width=\linewidth]{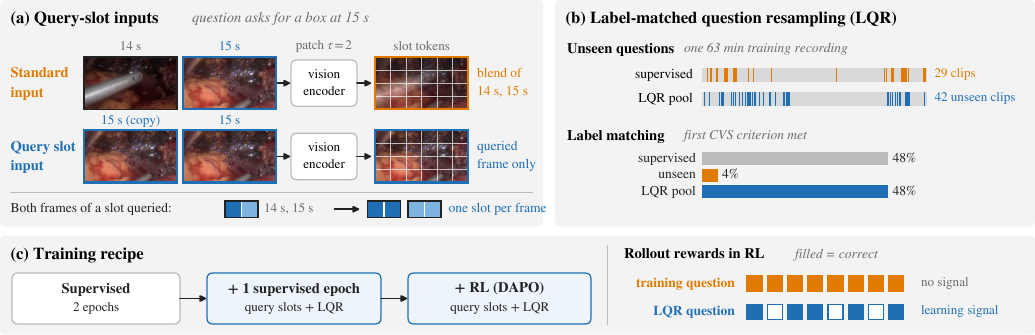}
\caption{\textbf{Evidence-aligned learning.} (a) Query-slot inputs fill the queried frame's token slot with that frame; a slot with two queried frames is split. (b) \method draws unseen questions from the training recordings and matches their label distribution. (c) Both enter the final supervised epoch and reinforcement learning.}
\label{fig:overview}
\end{figure*}

The benchmark measures whether a score depends on the video. Evidence-aligned learning does the same for training (\cref{fig:overview}), with two components. \emph{Query-slot inputs} make the visual tokens match the frame a question asks about (\cref{sec:qslot}). \emph{Label-matched question resampling} (\method) supplies questions the model has not memorized (\cref{sec:fresh}). Both are used in a final supervised epoch and in reinforcement learning (RL) whose rewards are the benchmark metrics (\cref{sec:rl}).

\subsection{Query-slot inputs}
\label{sec:qslot}
Many video encoders split a clip into pairs of consecutive frames and embed each pair into one set of visual tokens, which we call a slot; this is the case for the Qwen2.5-VL, Qwen3-VL, Qwen3.5, and GLM-4.1V/4.5V encoders~\cite{bai2025qwen25vl,bai2025qwen3vl,glmv2025glm41v,qwen2026qwen35}. The two frames of a pair share every token, and the model is told only their mean timestamp. A grounding question asks for a box at a stated time, yet the frame at that time usually shares its tokens with a neighbor. The effect is strongest when frames are far apart, as in PhaKIR, where grounding asks about the last of 16 frames sampled one second apart.

Query-slot inputs remove the blend. For each queried frame, we replace the other frame of its slot with a copy of the queried frame. The slot then encodes the queried frame alone, exactly as the encoder treats a still image, and the number of frames, the timestamps, the prompt, and the number of tokens stay the same. When both frames of a slot are queried, we split it into two slots, each holding one of the frames twice under its own timestamp; this adds one slot per split and affects 13\% of grounding questions. The rule uses only the times stated in the question, never the answer, and the model still receives the whole clip, so it applies in supervised training, RL, and inference.

We apply it to every question that is answered from named frames: grounding, frame-level CVS, and scene graphs. Questions about motion keep their frame pairs, because the pair is what shows the motion.

Independently of slots, we give grounding questions twice the pixel budget per frame. The analysis in \cref{sec:experiments} separates the two changes and shows that extra pixels help mainly once frames are aligned.

\subsection{Label-matched question resampling}
\label{sec:fresh}
After supervised fine-tuning, the training questions have little left to teach. When we sample eight answers per question, the model answers 95\% of its training multiple-choice questions correctly, against 76\% of unseen questions about the same recordings and 71\% on validation recordings. This matters for RL in particular. The methods we build on learn from the differences among several sampled answers to the same question. If all of them earn the same reward, the question provides no gradient. Only 14\% of training questions produce both correct and incorrect answers, against 28\% of unseen questions (39\% and 60\% for set-valued tasks). DAPO skips uninformative questions but cannot create informative ones; \method does, by supplying new questions.

\noindent\textbf{Unseen questions without new annotation.}
\method draws on two sources, both within the training recordings, tasks, and label vocabulary.
\emph{Unsampled questions} were generated from the training recordings but not sampled into the supervised training set.
\emph{Re-windowed clips} apply the benchmark's question generators at new time offsets, which gives new clips, timestamps, and answers from the same annotations.
The resulting pool has the same size as the training set, 77,261 questions, of which 41\% are unseen.

\noindent\textbf{Matching the label distribution.}
Unseen questions do not have the answer statistics of the training set. At Endoscapes, the first CVS criterion is met in 4\% of unseen frames but 48\% of training frames; unseen Cholec80 instrument questions never have an empty answer, against 4.9\% in training. A model absorbs the answer statistics of the questions it trains on, whether by imitation or by reward. We therefore let unseen questions supply the new learning signal while the answer statistics stay those of the training set. \method keeps the number of questions for every task and dataset. Within CVS and temporal tasks it also keeps the number of questions per answer class: each combination of CVS criteria, and whether the event occurs. Each class is filled with unseen questions first and with training questions of the same class when unseen ones run out. For the supervised epoch, the pool also keeps the distribution of answer-set sizes (\cref{sec:supp_rl}).

\subsection{Training}
\label{sec:rl}

\noindent\textbf{Supervised stages and baseline.}
Training has three stages. We first fine-tune the language-model weights of Qwen3.5-4B~\cite{qwen2026qwen35} for two epochs on \bench, with the vision encoder frozen, and keep the checkpoint with the lowest validation loss. A third supervised epoch continues from it with query-slot inputs and \method questions, and RL follows. The \emph{SFT baseline} instead runs the third epoch with the original questions and inputs, so every comparison holds the amount of supervised training fixed at three epochs. A 27B model~\cite{qwen2026qwen38} follows the same stages. Training settings are in \cref{sec:supp_rl}.

\noindent\textbf{Reinforcement learning with benchmark rewards.}
Each reward lies in $[0,1]$ and is computed by the benchmark's own parser and metric for the task: exact match for choices, F1 for sets, edit similarity for ordered steps, interval F1 for temporal answers, mean IoU for boxes, and the fraction of correct criteria for CVS. Captions are scored against their references by a rubric-based judge model, and timed captions average interval F1 and caption quality, with extra predicted segments scored as zero. An answer that cannot be parsed earns zero. Because the rewards are the audited metrics, the shortcuts removed in \cref{sec:shortcuts} earn no reward.

We optimize with DAPO~\cite{yu2025dapo}, sampling eight answers per question and skipping questions whose answers all earn the same reward. Each step uses 48 questions from the \method pool with query-slot inputs, and training runs for 200 steps. To offset class imbalance in multiple-choice and CVS tasks, answers to questions with rare reference classes receive a larger weight in the update (\cref{sec:supp_rl}). MedGRPO~\cite{su2026medgrpo}, the closest prior recipe, instead rescales rewards by percentiles of the supervised model, adds a format bonus, and trains on the original questions.

\section{Experiments}
\label{sec:experiments}

\begin{table*}[t]
\centering
\caption{\textbf{Main results} ($\times100$) on \bench test (15,101 questions), blind windows (8,292; \cref{tab:temporal}), and unseen hospitals (HC HeiChole, Str Strasbourg, KU KUMC, PK PhaKIR; \cref{tab:ood}). QS: query slots; LQR: question resampling. Bold: block best; \mbox{\colorbox{green!18}{green} / \colorbox{red!15}{red}}: significantly above / below the SFT baseline (no cell is red); brackets: better row 95\% interval.}
\label{tab:main}
\footnotesize
\setlength{\tabcolsep}{2.7pt}
\newcommand{\ci}[1]{{\fontsize{5}{6}\selectfont #1}}
\resizebox{\textwidth}{!}{%
\begin{tabular}{@{}l ccccccccc ccc ccccc@{}}
\toprule
 & \multicolumn{9}{c}{In-distribution} & \multicolumn{3}{c}{Blind windows} & \multicolumn{5}{c}{OOD} \\
\cmidrule(lr){2-10}\cmidrule(lr){11-13}\cmidrule(l){14-18}
 & Temp. & Spat. & Choice & Set & Ord. & Comp. & Traj. & T-cap & Caption & Pres. & Loc. & Bound. & Ch.$_\mathrm{HC}$ & Set$_\mathrm{HC}$ & Set$_\mathrm{Str}$ & Sp.$_\mathrm{KU}$ & Sp.$_\mathrm{PK}$ \\
\midrule
\multicolumn{18}{@{}l}{\emph{Answers that ignore the video}} \\
Always ``none'' & 35.9 & 0.0 & 0.0 & 7.6 & 0.0 & 0.0 & 0.0 & 0.0 & -- & -- & -- & -- & -- & -- & -- & -- & -- \\
Most frequent answer & 35.9 & 5.9 & 22.7 & 37.6 & 5.1 & 44.4 & 50.0 & 25.3 & -- & -- & -- & -- & -- & -- & -- & -- & -- \\
Final model, no frames & 33.4 & 6.7 & 28.2 & 29.5 & 32.6 & 44.4 & 50.0 & 50.7 & -- & -- & -- & -- & -- & -- & -- & -- & -- \\
Best video-free answer & -- & -- & -- & -- & -- & -- & -- & -- & -- & 50.0 & 0.0 & 36.5 & 31.0 & 17.5 & 48.5 & 3.2 & 17.4 \\
\midrule
\multicolumn{18}{@{}l}{\emph{Reference models (one-shot)}} \\
InternVL3.5-8B~\cite{wang2025internvl3_5} & 16.6 & 7.1 & 26.1 & 24.5 & 5.1 & 34.5 & 31.9 & 6.1 & 46.8 & 53.9 & 14.1 & 12.6 & 22.7 & 40.0 & 27.6 & 14.2 & 13.6 \\
Hulu-Med-7B~\cite{jiang2025hulumed} & 33.5 & 4.1 & 23.3 & 27.3 & 1.5 & 35.9 & 27.7 & 1.7 & 48.4 & 50.3 & 0.0 & 0.0 & 26.8 & 50.5 & 48.7 & 0.2 & 15.9 \\
MedGRPO-4B~\cite{su2026medgrpo} & 18.5 & 21.0 & 26.2 & 29.5 & 5.5 & 46.0 & 29.7 & 50.7 & 52.0 & 61.1 & 34.2 & 46.6 & 18.2 & 50.6 & 43.1 & 23.6 & 30.1 \\
Qwen3.5-4B~\cite{qwen2026qwen35} & 35.6 & 6.6 & 26.7 & 23.6 & 9.5 & 36.5 & 29.3 & 11.9 & 54.1 & 67.5 & 24.2 & 21.0 & 15.6 & 53.6 & 39.7 & 9.5 & 8.8 \\
Qwen3.8-27B~\cite{qwen2026qwen38} & 17.4 & 20.3 & 29.2 & 32.9 & 16.9 & 45.0 & 43.3 & 37.8 & 53.5 & 56.3 & 40.0 & 34.7 & 21.3 & 53.8 & 44.2 & 26.6 & 27.0 \\
Gemini 3.8 Flash~\cite{google2026gemini38flash} & 32.8 & 9.7 & 22.7 & 29.8 & 21.8 & 42.4 & 32.2 & 35.3 & 58.9 & 53.4 & 15.8 & 19.9 & 26.0 & 41.5 & 41.4 & 17.9 & 13.9 \\
Gemini 4 Argon~\cite{google2026gemini4argon} & 32.4 & 14.9 & 24.8 & 31.5 & 21.3 & 40.1 & 50.0 & 38.2 & 53.8 & 53.3 & 7.6 & 6.8 & 22.8 & 39.8 & 38.0 & 11.7 & 21.4 \\
GPT-6 Luna~\cite{openai2026gpt6luna} & 26.7 & 23.8 & 38.0 & 39.9 & 34.1 & 50.9 & 44.3 & 41.3 & 60.6 & 69.2 & 46.8 & 45.5 & 48.9 & 56.1 & 53.1 & 36.8 & 37.0 \\
\midrule
\multicolumn{18}{@{}l}{\emph{Qwen3.5-4B trained on \bench}} \\
SFT baseline & 53.3 & 30.4 & \best{70.4} & 68.8 & 78.6 & 55.5 & 48.8 & 70.1 & 73.9 & 81.2 & 49.6 & 59.0 & 61.6 & \best{59.8} & 63.7 & 33.3 & 33.9 \\ %
\quad + QS, LQR & \gain \best{55.2} & \gain 42.0 & 69.7 & \best{70.3} & \best{78.9} & \best{57.8} & 48.1 & 70.7 & 73.9 & \gain 82.7 & 50.3 & 59.5 & 62.4 & 59.2 & 67.3 & 33.5 & \gain 72.2 \\
\quad\quad + RL & 54.0 & \gain \best{44.4} & 70.1 & 69.7 & 78.2 & 57.1 & \best{51.1} & \gain \best{72.7} & \gain \best{75.0} & \gain \best{86.7} & \gain \best{57.7} & \gain \best{65.3} & \best{63.5} & 59.1 & \best{67.5} & \gain \best{38.9} & \gain \best{72.5} \\[-3pt]
\ci{\quad\quad gain over baseline, 95\% interval} & \ci{[0.5,3.1]} & \ci{[13.0,14.8]} & \ci{[$-$1.3,0.9]} & \ci{[0.0,2.9]} & \ci{[$-$1.9,2.4]} & \ci{[$-$0.2,4.7]} & \ci{[$-$3.1,7.3]} & \ci{[1.4,3.8]} & \ci{[0.1,2.6]} & \ci{[1.1,10.5]} & \ci{[3.9,12.8]} & \ci{[2.7,11.8]} & \ci{[0.0,3.9]} & \ci{[$-$2.6,1.3]} & \ci{[$-$0.7,8.6]} & \ci{[3.1,8.0]} & \ci{[36.1,42.2]} \\ %
\midrule
\multicolumn{18}{@{}l}{\emph{Qwen3.8-27B trained on \bench}} \\
SFT baseline & 55.8 & 42.7 & \best{71.8} & 70.4 & 77.3 & 54.4 & \best{48.2} & 72.1 & 74.0 & 83.1 & 56.2 & \best{61.4} & 63.0 & \best{58.0} & 65.7 & 42.3 & 42.0 \\ %
\quad + QS, LQR & \gain \best{57.9} & \gain 53.1 & 71.6 & 71.1 & 78.7 & \gain \best{57.1} & 47.9 & \gain 73.5 & \best{74.8} & 84.2 & 54.1 & 57.5 & 63.8 & 56.4 & 70.6 & \gain 49.7 & \gain 75.0 \\
\quad\quad + RL & 56.6 & \gain \best{53.9} & \best{71.8} & \best{71.8} & \gain \best{79.7} & \gain 56.8 & 46.0 & \gain \best{74.7} & 74.4 & \gain \best{87.8} & \best{59.3} & 59.2 & \best{63.9} & 56.1 & \best{70.7} & \gain \best{50.8} & \gain \best{75.2} \\[-3pt]
\ci{\quad\quad gain over baseline, 95\% interval} & \ci{[0.6,3.5]} & \ci{[10.1,12.1]} & \ci{[$-$1.1,1.0]} & \ci{[$-$0.2,3.2]} & \ci{[0.2,5.0]} & \ci{[0.7,4.7]} & \ci{[$-$4.2,4.3]} & \ci{[1.6,3.6]} & \ci{[$-$0.4,2.3]} & \ci{[2.4,7.4]} & \ci{[$-$0.4,6.0]} & \ci{[$-$7.4,2.4]} & \ci{[$-$1.2,3.0]} & \ci{[$-$3.7,0.2]} & \ci{[$-$0.8,10.1]} & \ci{[5.6,11.3]} & \ci{[28.6,40.7]} \\ %
\bottomrule
\end{tabular}}
\end{table*}

\begin{table}[t]
\centering
\caption{\textbf{Contribution of each component} (4B, $\times100$). Middle: third-epoch variants; bottom: RL (ours or MedGRPO's~\cite{su2026medgrpo}). QS: query slots; px: grounding pixels; KU, PK: KUMC, PhaKIR. Bold: best; \colorbox{green!18}{green} / \colorbox{red!15}{red}: significantly above / below baseline.}
\label{tab:ablation}
\scriptsize
\setlength{\tabcolsep}{1.0pt}
\begin{tabular*}{\linewidth}{@{\extracolsep{\fill}}l cccccc cc cc@{}}
\toprule
 & \multicolumn{6}{c}{Curated} & \multicolumn{2}{c}{Blind} & \multicolumn{2}{c}{OOD Gr.} \\
\cmidrule(lr){2-7}\cmidrule(lr){8-9}\cmidrule(l){10-11}
 & Temp. & Spat. & Ch. & Set & Comp. & T-cap & Pres. & Loc. & KU & PK \\
\midrule
SFT baseline & 53.3 & 30.4 & 70.4 & 68.8 & 55.5 & 70.1 & 81.2 & 49.6 & 33.3 & 33.9 \\
\midrule
2$\times$ px & 53.5 & \gain 31.5 & 70.6 & 68.5 & 54.5 & 70.1 & 80.9 & 49.8 & 33.1 & 33.7 \\
QS at inference & 53.2 & \gain 33.7 & 70.4 & 68.4 & 55.3 & 70.0 & 81.2 & 50.5 & 34.7 & \gain 54.7 \\
QS, 1$\times$ px & 53.1 & \gain 36.8 & 70.5 & 68.9 & \loss 54.2 & 70.4 & 81.0 & \gain 51.1 & 34.0 & \gain 62.9 \\
QS & 52.9 & \gain 41.2 & 70.7 & 69.5 & 56.0 & 69.9 & 81.2 & 50.3 & 34.8 & \gain 70.2 \\
QS, LQR (ours) & \gain \best{55.2} & \gain 42.0 & 69.7 & \best{70.3} & \best{57.8} & 70.7 & \gain 82.7 & 50.3 & 33.5 & \gain 72.2 \\
\midrule
Baseline + MedGRPO & 53.4 & \gain 32.5 & 70.4 & 68.5 & \loss 53.9 & \gain 72.2 & \gain 85.8 & \gain 56.3 & \gain 36.2 & \gain 35.5 \\
Baseline + RL & 53.3 & \gain 33.1 & \gain \best{71.1} & 69.5 & 54.2 & \gain 72.3 & \gain \best{86.9} & \gain \best{59.1} & \gain 38.0 & \gain 36.3 \\
QS, LQR + RL (ours) & 54.0 & \gain \best{44.4} & 70.1 & 69.7 & 57.1 & \gain \best{72.7} & \gain 86.7 & \gain 57.7 & \gain \best{38.9} & \gain \best{72.5} \\
\bottomrule
\end{tabular*}
\end{table}

\begin{table}[t]
\centering
\caption{\textbf{Temporal presence and localization} ($\times100$). F1: usual temporal F1; Pres.: mean of event recall and empty accuracy; Loc.: F1 on event clips; Bound.: windows with a boundary. Bold: block best; \colorbox{green!18}{green} / \colorbox{red!15}{red}: significantly above / below baseline.}
\label{tab:temporal}\label{tab:realistic}
\scriptsize
\setlength{\tabcolsep}{1.2pt}
\begin{tabular*}{\linewidth}{@{\extracolsep{\fill}}l ccccc ccc@{}}
\toprule
 & \multicolumn{5}{c}{Curated test} & \multicolumn{3}{c}{Blind windows} \\
\cmidrule(lr){2-6}\cmidrule(l){7-9}
Model & F1 & Pres. & Recall & Empty & Loc. & Pres. & Loc. & Bound. \\
\midrule
Video-free answer & 35.9 & 50.0 & 0.0 & 100.0 & 0.0 & 50.0 & 0.0 & 36.5 \\
\midrule
InternVL3.5-8B & 16.6 & 55.3 & 78.1 & 32.5 & 5.1 & 53.9 & 14.1 & 12.6 \\
Hulu-Med-7B & 33.5 & 48.3 & 4.9 & 91.7 & 0.6 & 50.3 & 0.0 & 0.0 \\
MedGRPO-4B & 18.5 & 53.2 & 87.5 & 18.9 & 19.2 & 61.1 & 34.2 & 46.6 \\
Qwen3.5-4B & 35.6 & 53.8 & 33.9 & 73.8 & 11.1 & 67.5 & 24.2 & 21.0 \\
Qwen3.8-27B & 17.4 & 52.7 & 98.9 & 6.4 & 24.4 & 56.3 & 40.0 & 34.7 \\
Gemini 3.8 Flash & 32.8 & 50.6 & 24.8 & 76.4 & 6.1 & 53.4 & 15.8 & 19.9 \\
Gemini 4 Argon & 32.4 & 52.9 & 28.4 & 77.4 & 7.0 & 53.3 & 7.6 & 6.8 \\
\midrule
SFT baseline, 4B & 53.3 & \best{83.4} & 82.1 & 84.6 & 36.1 & 81.2 & 49.6 & 59.0 \\
\quad + QS, LQR & \gain \best{55.2} & 82.8 & \loss 78.7 & \gain \best{86.9} & 36.7 & \gain 82.7 & 50.3 & 59.5 \\
\quad\quad + RL & 54.0 & 82.1 & \gain \best{87.9} & \loss 76.3 & \gain \best{40.8} & \gain \best{86.7} & \gain \best{57.7} & \gain \best{65.3} \\
SFT baseline, 27B & 55.8 & 83.0 & 82.2 & 83.8 & 39.3 & 83.1 & 56.2 & \best{61.4} \\
\quad + QS, LQR & \gain \best{57.9} & \best{83.2} & \loss 78.0 & \gain \best{88.4} & 39.4 & 84.2 & 54.1 & 57.5 \\
\quad\quad + RL & 56.6 & 83.0 & \gain \best{86.3} & \loss 79.8 & \gain \best{42.3} & \gain \best{87.8} & \best{59.3} & 59.2 \\
\bottomrule
\end{tabular*}
\end{table}

\noindent\textbf{Setup.}
All models answer the same questions and are scored with the family metrics of \cref{tab:families}. A judge model rates caption quality, and an unparseable answer scores 0. A family score averages datasets within each task and then tasks within the family. For paired comparisons we resample whole recordings 1,000 times, which respects the dependence among questions from the same recording, and call a difference significant when its 95\% interval excludes zero; brackets give these intervals. Running the same checkpoint twice moves family scores by about one point (\cref{sec:supp_noise}), so we treat smaller differences as ties.

We compare with three groups of models. \emph{Reference models} have not been trained on \bench and receive the task prompt with one worked example: open video models (InternVL3.5-8B~\cite{wang2025internvl3_5}, the medical generalist Hulu-Med-7B~\cite{jiang2025hulumed}, and the Qwen models we fine-tune) and commercial models (Gemini and GPT). \emph{Prior work} is the released MedGRPO model~\cite{su2026medgrpo} and, on its original test, our backbone trained on MedVidBench. \emph{Video-free baselines} answer without frames and show how much of each score needs the video. Evaluation details are in \cref{sec:supp_eval}.

\subsection{Main results}
\begin{figure*}[t]
\centering
\includegraphics[width=\linewidth]{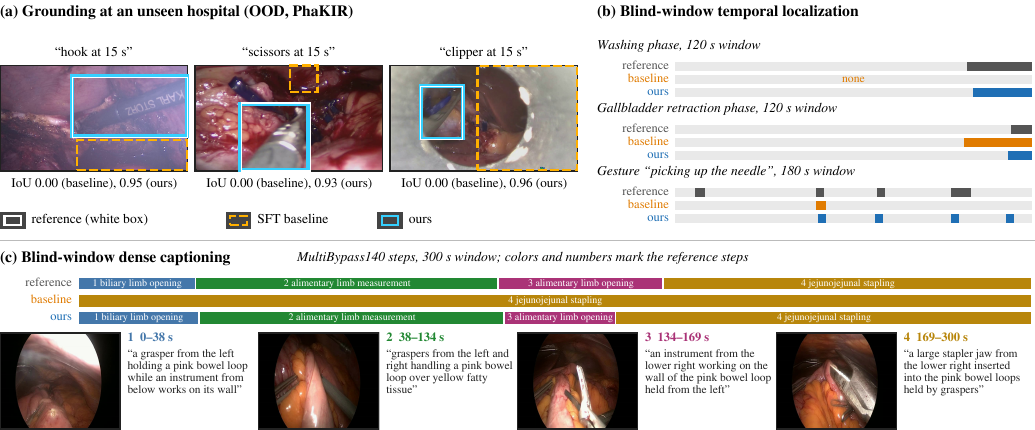}
\caption{\textbf{Qualitative results.} (a) PhaKIR frames from an unseen hospital (OOD): the final 4B model boxes the instrument at the queried time. (b) Blind windows: it returns intervals where the SFT baseline answers \texttt{none}, extends past the event, or finds one of several occurrences. (c) Dense captioning of a blind window: it separates and describes four steps where the baseline returns one.}
\label{fig:qualitative}
\end{figure*}

\noindent\textbf{Training on \bench.}
The SFT baseline exceeds the released MedGRPO model in every family (\cref{tab:main}), for example 68.8 against 29.5 set F1 and 78.6 against 5.5 step-sequence similarity. It is also competitive on the original MedVidBench test (\cref{tab:medvidbench_test}), whose prompts it never saw; there the final 4B model has the best grounding IoU, 31.8 against 24.4 for the backbone trained on MedVidBench.

\noindent\textbf{What the metrics expose.}
Without frames, the final 4B model loses more than half of its spatial, choice, set, and ordered scores. Temporal and timed-caption scores fall less, to 33.4 and 50.7, because empty answers and regular clip structure can be guessed; this is what the presence and localization scores and the blind windows are designed to expose (\cref{sec:shortcuts}). \cref{tab:temporal} shows the effect on the curated test. Under the usual temporal F1, the video-free answer (35.9) matches or exceeds every one-shot model, and Hulu-Med-7B, which mostly answers \texttt{none}, outscores Qwen3.8-27B (33.5 against 17.4), the best one-shot model at localization. Presence places every one-shot model near 50 and every trained model above 80. Models rank similarly on curated clips and blind windows (Spearman $\rho$ of 0.85 for presence and 0.98 for localization). CVS trajectories remain open: no model clearly exceeds the constant-answer score of 50.

\noindent\textbf{Evidence-aligned learning.}
Evidence-aligned learning improves the answers that depend on finding the right frame or region without a significant change in recognition (\cref{tab:main}). Compared with the SFT baseline, which has the same amount of supervised training, it raises grounding IoU from 30.4 to 44.4 ([13.0, 14.8]), timed-caption F1 by 2.6 ([1.4, 3.8]), and caption quality from 73.9 to 75.0. On blind windows, balanced presence rises from 81.2 to 86.7 ([1.1, 10.5]) and localization F1 from 49.6 to 57.7 ([3.9, 12.8]) (\cref{tab:realistic}); \cref{fig:qualitative} shows examples. Choice, set, ordered, and component scores stay within two points of the baseline, with no significant change, so the better localization does not come at the cost of recognition.

\noindent\textbf{Comparison with a larger model and with MedGRPO.}
The final 4B model exceeds the 27B SFT baseline in grounding (44.4 against 42.7, [0.4, 2.8]) and in blind-window presence (86.7 against 83.1, [0.4, 7.0]), and matches its localization (57.7 against 56.2). Compared with the MedGRPO recipe applied to the 4B SFT baseline, it gains 11.8 grounding points ([10.9, 12.7]) and 3.2 component-accuracy points ([0.9, 5.5]), and raises PhaKIR grounding from 35.5 to 72.5. On blind windows it scores 86.7 presence and 57.7 localization, against 85.8 and 56.3.

\subsection{Component analysis}
\cref{tab:ablation} adds components one at a time from one checkpoint.

\noindent\textbf{Query-slot inputs.}
At the standard pixel budget, query-slot inputs improve grounding over the baseline by 6.4 points ([5.6, 7.1]) and PhaKIR instrument IoU by 29.0 ([26.0, 32.1]). Part of the gain needs no training: using them only at inference adds 3.3 ([2.6, 3.8]) and 20.8 ([16.8, 26.2]). Resolution alone does little. Doubling the pixel budget with standard inputs adds 1.0 grounding IoU ([0.4, 1.6]) and nothing on PhaKIR, whereas with slots the same doubling adds 2.1 and 7.3. Splitting slots that hold two queried frames and extending the rule to frame-level CVS and scene-graph questions add a further 2.3 grounding points ([1.6, 2.9]), for a total of 10.7 over the baseline ([9.8, 11.6]). On the 13\% of grounding questions with a split slot, IoU rises from 46.7 to 58.9, and frame-level CVS accuracy rises from 66.3 to 70.5. The gain therefore comes from showing the model the frame the question asks about, not from more pixels. Conversely, a model trained with slots but evaluated on standard inputs keeps only 0.9 points of its grounding gain and 2.4 PhaKIR points (\cref{sec:supp_qs_crossover}). The same holds after RL: with standard inputs, RL from the baseline stays 11.3 grounding and 36.2 PhaKIR points below the full recipe.

\begin{table}[t]
\centering
\caption{\textbf{Queried frames without the clip} (grounding IoU, 4B). Full clip: standard evaluation; Frames: only the queried frames, matched frame size, same question. Bold: better input.}
\label{tab:frame_only}
\scriptsize
\setlength{\tabcolsep}{0pt}
\begin{tabular*}{\columnwidth}{@{\extracolsep{\fill}}l cc cc cc@{}}
\toprule
 & \multicolumn{2}{c}{Curated} & \multicolumn{2}{c}{KUMC} & \multicolumn{2}{c}{PhaKIR} \\
\cmidrule(lr){2-3}\cmidrule(lr){4-5}\cmidrule(lr){6-7}
Model & Full clip & Frames & Full clip & Frames & Full clip & Frames \\
\midrule
SFT baseline & 30.5 & \best{32.7} & \best{33.3} & 32.5 & 33.9 & \best{54.2} \\
\quad + QS & \best{41.5} & 40.0 & \best{34.8} & 33.9 & 70.2 & \best{71.3} \\
\quad\quad + LQR, RL (final) & \best{44.5} & 42.7 & \best{38.9} & 38.6 & \best{72.5} & 71.9 \\
\bottomrule
\end{tabular*}
\vspace{-4.5mm}
\end{table}

\noindent\textbf{Queried frames without the clip.}
\cref{tab:frame_only} asks whether a video model with query slots grounds as well as one that receives the queried frame as an image: we replace the clip with only the queried frames, at matched frame size (\cref{sec:supp_frame_only}). The SFT baseline improves when the other frames are removed, by 2.2 points on the curated test ([1.3, 3.1]) and 20.3 on PhaKIR ([15.1, 27.8]): its standard video input hides the queried frame. With query slots, the model reads the queried frame from the clip as well as from an image at both unseen hospitals, and the clip adds 1.5 points on the curated test ([0.7, 2.3]) and 1.8 for the final model ([1.0, 2.6]). The baseline with frame-only input stays far below the slot-trained model (32.7 against 41.5 curated, 54.2 against 70.2 on PhaKIR), so the slot-trained video model exceeds what the baseline reaches with the queried frame alone, and it keeps the surrounding video.

\noindent\textbf{\method in supervised training.}
Replacing the original questions with the \method pool in the third epoch adds 2.3 temporal F1 ([0.9, 3.7]) and 2.1 PhaKIR grounding points ([0.2, 3.9]), and raises blind-window presence from 81.2 to 82.7 ([0.2, 3.2]), with no new annotation.

\noindent\textbf{Reinforcement learning.}
RL adds 2.4 grounding points ([1.9, 2.9]) and 2.0 timed-caption F1 ([1.2, 3.0]), and improves KUMC grounding from 33.5 to 38.9 ([3.4, 7.8]). Its largest effect is on untrimmed video. On blind windows it adds 3.9 presence points ([0.7, 7.7]) and 7.4 localization F1 ([4.9, 10.0]), and boundary F1 rises from 59.5 to 65.3.
On curated clips, RL makes the model more willing to predict an event: event recall rises from 78.7 to 87.9 while accuracy on empty clips falls from 86.9 to 76.3, and localization F1 rises from 36.7 to 40.8 (\cref{tab:temporal}). A single temporal F1 hides this trade; the separate presence and localization scores show it.
With the class weight, RL keeps CVS component accuracy (57.1 against 57.8).

\subsection{Unseen hospitals and model scale}

\noindent\textbf{Unseen hospitals.}
Supervised training on \bench transfers to hospitals that contribute no training data (\cref{tab:main}; all columns in \cref{tab:ood}). Compared with the released MedGRPO model, HeiChole choice accuracy rises from 18.2 to 61.6, Strasbourg set F1 from 43.1 to 63.7, and KUMC grounding from 23.6 to 33.3.
The full recipe then raises PhaKIR grounding from 33.9 to 72.5, well above the 27B SFT baseline's 42.0, and KUMC grounding to 38.9. Strasbourg choice accuracy rises by 2.5 ([0.7, 4.7]); the other HeiChole and Strasbourg recognition and set scores do not change significantly. The commercial models trail the final 4B model in every column of \cref{tab:ood}; the strongest, GPT-6 Luna, reaches 37.0 grounding IoU on PhaKIR and 36.8 on KUMC.

\noindent\textbf{Scale.}
With standard inputs, the 27B SFT baseline leads the 4B one by 12.3 grounding points (42.7 against 30.4) and by at most 2.5 points in the other curated families, so grounding is where the larger model helps. Evidence-aligned learning also helps the larger model. Query-slot inputs and the \method epoch raise 27B grounding to 53.1 (+10.4, [9.4, 11.4]) and improve temporal, component, and timed-caption scores. RL then raises grounding to 53.9 and brings blind-window presence and localization to 87.8 and 59.3, from 83.1 and 56.2 for the 27B SFT baseline. At unseen hospitals, the 27B model reaches 75.2 grounding IoU on PhaKIR and 50.8 on KUMC, compared with 72.5 and 38.9 for the final 4B model.

\section{Conclusion}
\label{sec:conclusion}

A medical video language model can score well without using the evidence a question asks about, and the causes lie in its data, its visual input, and its training questions. \bench audits prompts and metrics with answers given without video, checks references against the frames, and tests on unseen hospitals and on windows cut blind to the events. Evidence-aligned learning then makes the visual tokens match the queried frame and trains on unseen questions whose answer statistics match the training set. Together they lift a 4B model above the grounding accuracy of a supervised 27B model and improve event detection and localization in untrimmed video. Query-slot inputs add almost no tokens, and \method needs no new annotation, so both apply to other models whose encoders merge frames and to other benchmarks built from dense annotations.

\noindent\textbf{Limitations and future work.}
Query-slot inputs apply when a question names its frames: grounding, frame-level CVS, and scene-graph questions here. Extending them to questions that leave the frame implicit, and adding recordings paired with clinical reports, are future work.

{
    \small
    \section*{LLM Usage Statement}
    Generative AI models (Anthropic Claude and OpenAI Codex) assisted this work in three ways: writing and debugging code, running and analysing experiments, and drafting and editing the text, tables, and figures of the paper and its supplement. The authors designed the benchmark, the method, and the experiments, checked every result and all text, and take full responsibility for the content.

    \bibliographystyle{ieeenat_fullname}
    \bibliography{main}

\begin{thebibliography}{47}
\providecommand{\natexlab}[1]{#1}
\providecommand{\url}[1]{\texttt{#1}}
\expandafter\ifx\csname urlstyle\endcsname\relax
  \providecommand{\doi}[1]{doi: #1}\else
  \providecommand{\doi}{doi: \begingroup \urlstyle{rm}\Url}\fi

\bibitem[Alabi et~al.(2025)]{alabi2025cholecinstanceseg}
Oluwatosin Alabi et~al.
\newblock {CholecInstanceSeg}: A tool instance segmentation dataset for
  laparoscopic surgery.
\newblock \emph{Scientific Data}, 12:\penalty0 825, 2025.

\bibitem[Arnab et~al.(2021)Arnab, Dehghani, Heigold, Sun, Lu{\v{c}}i{\'c}, and
  Schmid]{arnab2021vivit}
Anurag Arnab, Mostafa Dehghani, Georg Heigold, Chen Sun, Mario Lu{\v{c}}i{\'c},
  and Cordelia Schmid.
\newblock {ViViT}: A video vision transformer.
\newblock In \emph{ICCV}, 2021.

\bibitem[Bai et~al.(2025{\natexlab{a}})Bai, Cai, Chen, et~al.]{bai2025qwen3vl}
Shuai Bai, Yuxuan Cai, Ruizhe Chen, et~al.
\newblock {Qwen3-VL} technical report, 2025{\natexlab{a}}.

\bibitem[Bai et~al.(2025{\natexlab{b}})Bai, Chen, Liu, et~al.]{bai2025qwen25vl}
Shuai Bai, Keqin Chen, Xuejing Liu, et~al.
\newblock {Qwen2.5-VL} technical report, 2025{\natexlab{b}}.

\bibitem[Biffi et~al.(2024)]{biffi2024realcolon}
Carlo Biffi et~al.
\newblock {REAL-Colon}: A dataset for developing real-world {AI} applications
  in colonoscopy.
\newblock \emph{Scientific Data}, 11:\penalty0 539, 2024.

\bibitem[Buch et~al.(2022)Buch, Eyzaguirre, Gaidon, Wu, Fei-Fei, and
  Niebles]{buch2022revisiting}
Shyamal Buch, Crist{\'o}bal Eyzaguirre, Adrien Gaidon, Jiajun Wu, Li Fei-Fei,
  and Juan~Carlos Niebles.
\newblock Revisiting the ``video'' in video-language understanding.
\newblock In \emph{CVPR}, 2022.

\bibitem[Chen et~al.(2026)Chen, Li, Xu, Liu, Cui, Xu, Teoh, He, and
  Qin]{prostatd2025}
Yiliang Chen, Zhixi Li, Cheng Xu, Alex~Qinyang Liu, Ruize Cui, Xuemiao Xu,
  Jeremy Yuen-Chun Teoh, Shengfeng He, and Jing Qin.
\newblock {ProstaTD}: Bridging surgical triplet from classification to fully
  supervised detection.
\newblock In \emph{ICLR}, 2026.

\bibitem[Drago et~al.(2025)Drago, Carlini, Balyemez, Pierantozzi, Lena, Hassan,
  Stoyanov, De~Momi, Bano, and Hoque]{surgvivqa2025}
Mauro~Orazio Drago, Luca Carlini, Pelinsu~Celebi Balyemez, Dennis Pierantozzi,
  Chiara Lena, Cesare Hassan, Danail Stoyanov, Elena De~Momi, Sophia Bano, and
  Mobarak~I. Hoque.
\newblock {SurgViVQA}: Temporally-grounded video question answering for
  surgical scene understanding.
\newblock arXiv:2511.03325, 2025.

\bibitem[Feng et~al.(2025)]{feng2025videor1}
Kaituo Feng et~al.
\newblock {Video-R1}: Reinforcing video reasoning in {MLLMs}.
\newblock arXiv:2503.21776, 2025.

\bibitem[Fujii et~al.(2024)]{fujii2024egosurgery}
Ryo Fujii et~al.
\newblock {EgoSurgery-Phase}: A dataset of surgical phase recognition from
  egocentric open surgery videos.
\newblock In \emph{MICCAI}, 2024.

\bibitem[Gao et~al.(2014)Gao, Vedula, Reiley, Ahmidi, Varadarajan, Lin, Tao,
  Zappella, B{\'e}jar, Yuh, Chen, Vidal, Khudanpur, and Hager]{gao2014jigsaws}
Yixin Gao, S.~Swaroop Vedula, Carol~E. Reiley, Narges Ahmidi, Balakrishnan
  Varadarajan, Henry~C. Lin, Lingling Tao, Luca Zappella, Benjam{\'\i}n
  B{\'e}jar, David~D. Yuh, Chi Chiung~Grace Chen, Ren{\'e} Vidal, Sanjeev
  Khudanpur, and Gregory~D. Hager.
\newblock {JHU-ISI} gesture and skill assessment working set ({JIGSAWS}): A
  surgical activity dataset for human motion modeling.
\newblock In \emph{MICCAI Workshop: M2CAI}, 2014.

\bibitem[Geirhos et~al.(2020)Geirhos, Jacobsen, Michaelis, Zemel, Brendel,
  Bethge, and Wichmann]{geirhos2020shortcut}
Robert Geirhos, J{\"o}rn-Henrik Jacobsen, Claudio Michaelis, Richard Zemel,
  Wieland Brendel, Matthias Bethge, and Felix~A. Wichmann.
\newblock Shortcut learning in deep neural networks.
\newblock \emph{Nature Machine Intelligence}, 2\penalty0 (11):\penalty0
  665--673, 2020.

\bibitem[{GLM-V Team}(2025)]{glmv2025glm41v}
{GLM-V Team}.
\newblock {GLM-4.5V} and {GLM-4.1V-Thinking}: Towards versatile multimodal
  reasoning with scalable reinforcement learning, 2025.

\bibitem[Goodman et~al.(2021)]{goodman2021avos}
Emmett~D. Goodman et~al.
\newblock A real-time spatiotemporal {AI} model analyzes skill in open surgical
  videos.
\newblock arXiv:2112.07219, 2021.

\bibitem[{Google}(2026{\natexlab{a}})]{google2026gemini38flash}
{Google}.
\newblock {Gemini 3.8 Flash}, 2026{\natexlab{a}}.
\newblock Commercial model, accessed through the API in 2026.

\bibitem[{Google}(2026{\natexlab{b}})]{google2026gemini4argon}
{Google}.
\newblock {Gemini 4 Argon}, 2026{\natexlab{b}}.
\newblock Commercial model, accessed through the API in 2026.

\bibitem[Gupta et~al.(2022)Gupta, Attal, and Demner-Fushman]{gupta2022medvidqa}
Deepak Gupta, Kush Attal, and Dina Demner-Fushman.
\newblock A dataset for medical instructional video classification and question
  answering.
\newblock \emph{arXiv preprint arXiv:2201.12888}, 2022.

\bibitem[Hu et~al.(2023)]{hu2023nurvid}
Ming Hu et~al.
\newblock {NurViD}: A large expert-level video database for nursing procedure
  activity understanding.
\newblock In \emph{NeurIPS Datasets and Benchmarks Track}, 2023.

\bibitem[Huang et~al.(2026)Huang, Zou, Guo, Li, Ji, Wang, Wang, and
  Si]{huang2026surgmllm}
Jincai Huang, Shihao Zou, Yuchen Guo, Jingjing Li, Wei Ji, Kai Wang, Shanshan
  Wang, and Weixin Si.
\newblock Towards unified surgical scene understanding: Bridging reasoning and
  grounding via {MLLMs}.
\newblock In \emph{Medical Image Computing and Computer Assisted Intervention
  -- MICCAI 2026}. Springer Nature Switzerland, 2026.

\bibitem[Jiang et~al.(2025)Jiang, Wang, Song, Hu, Zhou, Pu, Zhang, Yang, Feng,
  Zhou, et~al.]{jiang2025hulumed}
Songtao Jiang, Yuan Wang, Sibo Song, Tianxiang Hu, Chenyi Zhou, Bin Pu, Yan
  Zhang, Zhibo Yang, Yang Feng, Joey~Tianyi Zhou, et~al.
\newblock {Hulu-Med}: A transparent generalist model towards holistic medical
  vision-language understanding.
\newblock \emph{arXiv preprint arXiv:2510.08668}, 2025.

\bibitem[Lavanchy et~al.(2024)]{lavanchy2024multibypass}
Jo{\"e}l~L. Lavanchy et~al.
\newblock Challenges in multi-centric generalization: phase and step
  recognition in {Roux-en-Y} gastric bypass surgery.
\newblock \emph{International Journal of Computer Assisted Radiology and
  Surgery}, 19\penalty0 (11):\penalty0 2249--2257, 2024.

\bibitem[Li et~al.(2024)Li, Skinner, Yang, Quaranto, Schwaitzberg, Kim, and
  Xiong]{li2024llavasurg}
Jiajie Li, Garrett Skinner, Gene Yang, Brian Quaranto, Steven Schwaitzberg,
  Peter Kim, and Jinjun Xiong.
\newblock {LLaVA-Surg}: Towards multimodal surgical assistant via structured
  surgical video learning.
\newblock \emph{arXiv preprint arXiv:2408.07981}, 2024.

\bibitem[Li et~al.(2021)Li, Fathan, Patel, et~al.]{li2021kumc}
Kaidong Li, Mohammad~I. Fathan, Krushi Patel, et~al.
\newblock Colonoscopy polyp detection and classification: Dataset creation and
  comparative evaluations.
\newblock \emph{PLoS ONE}, 16\penalty0 (8):\penalty0 e0255809, 2021.

\bibitem[Ma et~al.(2021)Ma, Chen, Cheng, Li, and Sun]{ma2021ldpolypvideo}
Yiting Ma, Xuejin Chen, Kai Cheng, Yang Li, and Bin Sun.
\newblock {LDPolypVideo} benchmark: A large-scale colonoscopy video dataset of
  diverse polyps.
\newblock In \emph{Medical Image Computing and Computer Assisted Intervention
  (MICCAI)}, 2021.

\bibitem[Mascagni et~al.(2025)Mascagni, Alapatt, Murali, Vardazaryan, Garcia,
  Okamoto, Costamagna, Mutter, Marescaux, Dallemagne, and
  Padoy]{murali2025endoscapes}
Pietro Mascagni, Deepak Alapatt, Aditya Murali, Armine Vardazaryan, Alain
  Garcia, Nariaki Okamoto, Guido Costamagna, Didier Mutter, Jacques Marescaux,
  Bernard Dallemagne, and Nicolas Padoy.
\newblock Endoscapes, a critical view of safety and surgical scene segmentation
  dataset for laparoscopic cholecystectomy.
\newblock \emph{Scientific Data}, 12:\penalty0 331, 2025.

\bibitem[Nwoye et~al.(2022)Nwoye, Yu, Gonzalez, Seeliger, Mascagni, Mutter,
  Marescaux, and Padoy]{nwoye2022rendezvous}
Chinedu~Innocent Nwoye, Tong Yu, Cristians Gonzalez, Barbara Seeliger, Pietro
  Mascagni, Didier Mutter, Jacques Marescaux, and Nicolas Padoy.
\newblock Rendezvous: Attention mechanisms for the recognition of surgical
  action triplets in endoscopic videos.
\newblock \emph{Medical Image Analysis}, 78:\penalty0 102433, 2022.

\bibitem[Nwoye et~al.(2025)]{nwoye2025cholectrack20}
Chinedu~Innocent Nwoye et~al.
\newblock {CholecTrack20}: A multi-perspective tracking dataset for surgical
  tools.
\newblock In \emph{CVPR}, 2025.

\bibitem[{OpenAI}(2026)]{openai2026gpt6luna}
{OpenAI}.
\newblock {GPT-6 Luna}, 2026.
\newblock Commercial model, accessed through the API in 2026.

\bibitem[{\"O}zsoy et~al.(2025)]{ozsoy2025egoexor}
Ege {\"O}zsoy et~al.
\newblock {EgoExOR}: An ego-exo-centric operating room dataset for surgical
  activity understanding.
\newblock In \emph{NeurIPS Datasets and Benchmarks Track}, 2025.

\bibitem[Perez et~al.(2026)Perez, Rau, Mlambo, White, Brunet, Verde, Caicedo,
  Lavanchy, Schmidt, Nwoye, Mohareri, and Jamal]{perez2026sureon}
Alejandra Perez, Anita Rau, Busisiwe Mlambo, Lee White, Aina Brunet, Juan
  Verde, Juan Caicedo, Jo{\"e}l~L. Lavanchy, Adam Schmidt, Chinedu Nwoye, Omid
  Mohareri, and Muhammad~Abdullah Jamal.
\newblock {SUREON}: A benchmark and vision-language model for surgical
  reasoning.
\newblock \emph{arXiv preprint arXiv:2603.06570}, 2026.

\bibitem[Psychogyios et~al.(2023)]{psychogyios2023sarrarp50}
Dimitrios Psychogyios et~al.
\newblock {SAR-RARP50}: Segmentation of surgical instrumentation and action
  recognition on robot-assisted radical prostatectomy challenge.
\newblock arXiv:2401.00496, 2023.

\bibitem[{Qwen Team}(2026{\natexlab{a}})]{qwen2026qwen35}
{Qwen Team}.
\newblock {Qwen3.5-4B}: Model card.
\newblock \url{https://huggingface.co/Qwen/Qwen3.5-4B}, 2026{\natexlab{a}}.
\newblock Accessed 17 September 2026.

\bibitem[{Qwen Team}(2026{\natexlab{b}})]{qwen2026qwen38}
{Qwen Team}.
\newblock {Qwen3.8}: Model card.
\newblock \url{https://huggingface.co/Qwen/Qwen3.8-27B}, 2026{\natexlab{b}}.
\newblock Accessed 17 November 2026.

\bibitem[R{\'\i}os et~al.(2023)]{rios2023cholec80cvs}
Manuel~Sebasti{\'a}n R{\'\i}os et~al.
\newblock {Cholec80-CVS}: An open dataset with an evaluation of {Strasberg's}
  critical view of safety for {AI}.
\newblock \emph{Scientific Data}, 10:\penalty0 194, 2023.

\bibitem[Rueckert et~al.(2025)]{rueckert2025phakir}
Tobias Rueckert et~al.
\newblock Video dataset for surgical phase, keypoint, and instrument
  recognition in laparoscopic surgery ({PhaKIR}), 2025.

\bibitem[Shao et~al.(2024)Shao, Wang, Zhu, Xu, Song, Bi, Zhang, Zhang, Li, Wu,
  and Guo]{shao2024deepseekmath}
Zhihong Shao, Peiyi Wang, Qihao Zhu, Runxin Xu, Junxiao Song, Xiao Bi, Haowei
  Zhang, Mingchuan Zhang, Y.~K. Li, Y. Wu, and Daya Guo.
\newblock {DeepSeekMath}: Pushing the limits of mathematical reasoning in open
  language models.
\newblock arXiv:2402.03300, 2024.

\bibitem[Su et~al.(2026)Su, Choudhuri, Gao, Planche, Nguyen, Zheng, Shen,
  Innanje, Chen, Elhamifar, and Wu]{su2026medgrpo}
Yuhao Su, Anwesa Choudhuri, Zhongpai Gao, Benjamin Planche, Van~Nguyen Nguyen,
  Meng Zheng, Yuhan Shen, Arun Innanje, Terrence Chen, Ehsan Elhamifar, and
  Ziyan Wu.
\newblock {MedGRPO}: Multi-task reinforcement learning for heterogeneous
  medical video understanding.
\newblock arXiv:2512.06581v4, 2026.
\newblock Accepted at CVPR 2026.

\bibitem[Tong et~al.(2022)Tong, Song, Wang, and Wang]{tong2022videomae}
Zhan Tong, Yibing Song, Jue Wang, and Limin Wang.
\newblock {VideoMAE}: Masked autoencoders are data-efficient learners for
  self-supervised video pre-training.
\newblock In \emph{NeurIPS}, 2022.

\bibitem[Twinanda et~al.(2016)Twinanda, Mutter, Marescaux, de~Mathelin, and
  Padoy]{twinanda2016m2cai}
Andru~P. Twinanda, Didier Mutter, Jacques Marescaux, Michel de Mathelin, and
  Nicolas Padoy.
\newblock Single- and multi-task architectures for surgical workflow challenge
  at {M2CAI} 2016.
\newblock arXiv:1610.08844, 2016.

\bibitem[Twinanda et~al.(2017)Twinanda, Shehata, Mutter, Marescaux,
  de~Mathelin, and Padoy]{twinanda2017endonet}
Andru~P. Twinanda, Sherif Shehata, Didier Mutter, Jacques Marescaux, Michel de
  Mathelin, and Nicolas Padoy.
\newblock {EndoNet}: A deep architecture for recognition tasks on laparoscopic
  videos.
\newblock \emph{IEEE Transactions on Medical Imaging}, 36\penalty0
  (1):\penalty0 86--97, 2017.

\bibitem[Wagner et~al.(2023)]{wagner2023heichole}
Martin Wagner et~al.
\newblock Comparative validation of machine learning algorithms for surgical
  workflow and skill analysis with the {HeiChole} benchmark.
\newblock \emph{Medical Image Analysis}, 86:\penalty0 102770, 2023.

\bibitem[Wang et~al.(2025{\natexlab{a}})Wang, Wang, Mo, Bai, Yuan, Hu, Wu, He,
  Huang, Padoy, Lei, Liu, Navab, and Ren]{wang2025surgvidlm}
Guankun Wang, Junyi Wang, Wenjin Mo, Long Bai, Kun Yuan, Ming Hu, Jinlin Wu,
  Junjun He, Yiming Huang, Nicolas Padoy, Zhen Lei, Hongbin Liu, Nassir Navab,
  and Hongliang Ren.
\newblock {SurgVidLM}: Towards multi-grained surgical video understanding with
  large language model.
\newblock \emph{arXiv preprint arXiv:2506.17873}, 2025{\natexlab{a}}.

\bibitem[Wang et~al.(2024)]{wang2024copesd}
Guankun Wang et~al.
\newblock {CoPESD}: A multi-level surgical motion dataset for training large
  vision-language models to co-pilot endoscopic submucosal dissection.
\newblock arXiv:2410.07540, 2024.

\bibitem[Wang et~al.(2025{\natexlab{b}})Wang, Gao, Gu, Pu, Cui, Wei, Liu, Jing,
  Ye, Shao, et~al.]{wang2025internvl3_5}
Weiyun Wang, Zhangwei Gao, Lixin Gu, Hengjun Pu, Long Cui, Xingguang Wei,
  Zhaoyang Liu, Linglin Jing, Shenglong Ye, Jie Shao, et~al.
\newblock {InternVL3.5}: Advancing open-source multimodal models in
  versatility, reasoning, and efficiency.
\newblock \emph{arXiv preprint arXiv:2508.18265}, 2025{\natexlab{b}}.

\bibitem[Wang et~al.(2022)]{wang2022autolaparo}
Ziyi Wang et~al.
\newblock {AutoLaparo}: A new dataset of integrated multi-tasks for
  image-guided surgical automation in laparoscopic hysterectomy.
\newblock In \emph{MICCAI}, 2022.

\bibitem[Yu et~al.(2025)]{yu2025dapo}
Qiying Yu et~al.
\newblock {DAPO}: An open-source {LLM} reinforcement learning system at scale.
\newblock arXiv:2503.14476, 2025.

\bibitem[Yu et~al.(2026)Yu, Zhao, Prudyus, Ruban, Yao, Yi, Grzybowski, and
  Jin]{yu2026lensnarrate}
Tao Yu, Kaikai Zhao, Vitalii Prudyus, Andrii Ruban, Rui Yao, Quanyong Yi,
  Andrzej Grzybowski, and Kai Jin.
\newblock Video-to-report generation for cataract surgery using procedurally
  grounded vision--language models.
\newblock \emph{Communications Medicine}, 2026.

\end{thebibliography}
}

\clearpage
\maketitlesupplementary
\appendix

\section{Benchmark details}
\label{sec:supp_data}
This section gives the composition of \bench, the corrections and sampling behind it, the caption audit, and the details of the test suites.

\subsection{Composition}
\begin{figure*}[t]
\centering
\includegraphics[width=\linewidth]{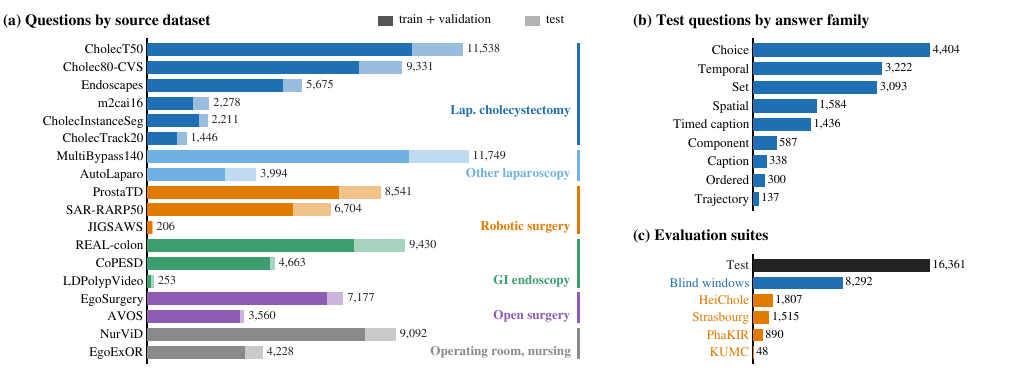}
\caption{\textbf{Composition of \bench.} (a) Questions per source dataset, colored by clinical domain. (b) Headline test questions per answer family. (c) Curated test, blind-window, and unseen-center suites.}
\label{fig:composition}
\end{figure*}

\Cref{tab:data_composition} lists the 18 source datasets with their procedures, task families, and recording-disjoint split sizes: 2,676 recordings and 102,076 questions in total. \Cref{fig:composition} shows the same composition by clinical domain, answer family, and evaluation suite. \Cref{tab:task_matrix} gives the question count of each of the 34 tasks in each dataset, and \cref{tab:task_matrix_ood} the counts in the blind-window and OOD suites.

\begin{table*}[t]
\centering
\caption{\textbf{Dataset composition of \bench.} The four Cholec* sources share one patient pool, and blind windows reuse test recordings; totals count unique recordings. Families: Te temporal, Sp spatial, Ch choice, Se set, Or ordered, Co component, Tr trajectory, TC timed caption, Ca caption.}
\label{tab:data_composition}
\footnotesize
\setlength{\tabcolsep}{3pt}
\begin{tabular*}{\textwidth}{@{\extracolsep{\fill}}l l c l rrr rrr@{}}
\toprule
 & & & & \multicolumn{3}{c}{Recordings} & \multicolumn{3}{c}{Questions} \\
\cmidrule(lr){5-7}\cmidrule(l){8-10}
Dataset & Procedure & Tasks & Families & Train & Val. & Test & Train & Val. & Test \\
\midrule
\multicolumn{10}{c}{\textbf{In-domain} (\bench train / validation / test splits)} \\
\midrule
AutoLaparo & lap.\ hysterectomy & 6 & Te, Sp, Ch, Se, TC & 180 & 44 & 80 & 2,311 & 549 & 1,134 \\
AVOS & open surgery (web video) & 3 & Te, Ch, TC & 31 & 4 & 8 & 3,267 & 112 & 181 \\
Cholec80-CVS & lap.\ cholecystectomy & 6 & Te, Ch, Se, Co, TC & 58 & 5 & 17 & 7,049 & 684 & 1,598 \\
CholecInstanceSeg & lap.\ cholecystectomy & 1 & Sp & 55 & 5 & 18 & 1,761 & 150 & 300 \\
CholecT50 & lap.\ cholecystectomy & 8 & Te, Ch, Se, TC, Ca & 32 & 4 & 14 & 8,581 & 1,102 & 1,855 \\
CholecTrack20 & lap.\ cholecystectomy & 2 & Sp, Ca & 11 & 1 & 6 & 917 & 163 & 366 \\
CoPESD & endoscopic submucosal dissection & 6 & Te, Sp, Ch, TC, Ca & 8 & 2 & 3 & 4,235 & 264 & 164 \\
EgoExOR & operating-room scenes (ego + exo) & 7 & Te, Se, TC, Ca & 22 & 8 & 10 & 3,175 & 396 & 657 \\
EgoSurgery & open surgery (egocentric) & 7 & Te, Sp, Ch, TC, Ca & 11 & 4 & 5 & 6,115 & 466 & 596 \\
Endoscapes & lap.\ cholecystectomy & 3 & Sp, Co, Tr & 151 & 10 & 40 & 4,603 & 348 & 724 \\
JIGSAWS & robotic bench-top training & 1 & Co & 78 & 9 & 16 & 156 & 18 & 32 \\
LDPolypVideo & colonoscopy & 1 & Sp & 69 & 19 & 50 & 120 & 40 & 93 \\
m2cai16 & lap.\ cholecystectomy & 4 & Te, Ch, TC & 24 & 3 & 14 & 1,531 & 166 & 581 \\
MultiBypass140 (Bern) & lap.\ gastric bypass & 9 & Te, Ch, Se, Or, TC & 42 & 10 & 18 & 8,447 & 1,117 & 2,185 \\
NurViD & nursing procedures & 4 & Te, Ch, TC & 1,038 & 153 & 322 & 7,362 & 588 & 1,142 \\
ProstaTD & robotic prostatectomy & 7 & Te, Se, TC, Ca & 11 & 2 & 8 & 6,339 & 670 & 1,532 \\
REAL-colon & colonoscopy & 10 & Te, Sp, Ch, Se, TC, Ca & 20 & 4 & 10 & 6,516 & 1,056 & 1,858 \\
SAR-RARP50 & robotic prostatectomy & 6 & Te, Sp, Ch, Se, TC & 36 & 4 & 10 & 4,776 & 565 & 1,363 \\
\midrule
\textbf{In-domain total} & & 34 & & \textbf{1,783} & \textbf{281} & \textbf{612} & \textbf{77,261} & \textbf{8,454} & \textbf{16,361} \\
\midrule[\heavyrulewidth]
\addlinespace[3pt]
\multicolumn{10}{c}{\textbf{Blind windows} (cut from in-domain test recordings; test only)} \\
\midrule
AutoLaparo & lap.\ hysterectomy & 4 & Te, Ch, TC & -- & -- & 7 & -- & -- & 967 \\
Cholec80-CVS & lap.\ cholecystectomy & 5 & Te, Ch, Se, TC & -- & -- & 17 & -- & -- & 1,373 \\
CholecT50 & lap.\ cholecystectomy & 5 & Te, Ch, Se, TC & -- & -- & 14 & -- & -- & 1,286 \\
EgoSurgery & open surgery (egocentric) & 2 & Te, Ch & -- & -- & 5 & -- & -- & 579 \\
m2cai16 & lap.\ cholecystectomy & 4 & Te, Ch, TC & -- & -- & 14 & -- & -- & 976 \\
MultiBypass140 (Bern) & lap.\ gastric bypass & 6 & Te, Ch, TC & -- & -- & 18 & -- & -- & 1,720 \\
ProstaTD & robotic prostatectomy & 3 & Te, Se & -- & -- & 8 & -- & -- & 734 \\
REAL-colon & colonoscopy & 1 & Te & -- & -- & 10 & -- & -- & 153 \\
SAR-RARP50 & robotic prostatectomy & 3 & Te, Se, TC & -- & -- & 10 & -- & -- & 504 \\
\midrule
\textbf{Blind-window total} & & 9 & & -- & -- & \textbf{90} & -- & -- & \textbf{8,292} \\
\midrule[\heavyrulewidth]
\addlinespace[3pt]
\multicolumn{10}{c}{\textbf{Out-of-distribution} (unseen hospitals; test only, no training data)} \\
\midrule
HeiChole & lap.\ cholecystectomy (2 hospitals) & 5 & Ch, Se, TC & -- & -- & 24 & -- & -- & 1,807 \\
MB140 Strasbourg & lap.\ gastric bypass & 7 & Ch, Se, Or, TC & -- & -- & 14 & -- & -- & 1,515 \\
KUMC & colonoscopy & 1 & Sp & -- & -- & 35 & -- & -- & 48 \\
PhaKIR & lap.\ cholecystectomy (3 centers) & 5 & Sp, Ch, Se, TC & -- & -- & 8 & -- & -- & 890 \\
\midrule
\textbf{OOD total} & & 11 & & -- & -- & \textbf{81} & -- & -- & \textbf{4,260} \\
\bottomrule
\end{tabular*}
\end{table*}

\begin{table*}[t]
\centering
\caption{\textbf{Questions per task and source dataset} (in-domain splits only: train + validation + test), grouped by answer family, with split totals per task. Empty cells: the dataset does not supply the task. Blind-window and OOD suites: \cref{tab:task_matrix_ood}.}
\label{tab:task_matrix}
\scriptsize
\setlength{\tabcolsep}{1.3pt}
\begin{tabular*}{\textwidth}{@{\extracolsep{\fill}}l rrrrrrrrrrrrrrrrrr rrr@{}}
\toprule
Task & \rotatebox{90}{CholecT50} & \rotatebox{90}{Cholec80-CVS} & \rotatebox{90}{Endoscapes} & \rotatebox{90}{m2cai16} & \rotatebox{90}{CholecInstanceSeg} & \rotatebox{90}{CholecTrack20} & \rotatebox{90}{MultiBypass140} & \rotatebox{90}{AutoLaparo} & \rotatebox{90}{ProstaTD} & \rotatebox{90}{SAR-RARP50} & \rotatebox{90}{JIGSAWS} & \rotatebox{90}{REAL-colon} & \rotatebox{90}{CoPESD} & \rotatebox{90}{LDPolypVideo} & \rotatebox{90}{EgoSurgery} & \rotatebox{90}{AVOS} & \rotatebox{90}{NurViD} & \rotatebox{90}{EgoExOR} & \rotatebox{90}{\textbf{Train}} & \rotatebox{90}{\textbf{Val.}} & \rotatebox{90}{\textbf{Test}} \\
\midrule
\multicolumn{22}{@{}l}{\emph{Temporal}} \\
\quad Action localization & 2,024 & 1,673 &  & 785 &  &  & 2,137 & 734 & 1,521 & 703 &  &  & 693 &  & 1,231 & 1,257 & 2,607 &  & 11,722 & 1,157 & 2,486 \\
\quad Instrument-action localization &  &  &  &  &  &  &  &  & 1,458 &  &  &  &  &  &  &  &  &  & 1,008 & 150 & 300 \\
\quad Activity localization &  &  &  &  &  &  &  &  &  &  &  &  &  &  &  &  &  & 1,701 & 1,251 & 150 & 300 \\
\quad Polyp visibility intervals &  &  &  &  &  &  &  &  &  &  &  & 648 &  &  &  &  &  &  & 467 & 20 & 161 \\
\midrule
\multicolumn{22}{@{}l}{\emph{Spatial}} \\
\quad Spatio-temporal grounding &  &  & 2,317 &  & 2,211 & 980 &  & 720 &  & 1,502 &  & 444 & 424 & 253 & 1,001 &  &  &  & 7,213 & 966 & 1,673 \\
\midrule
\multicolumn{22}{@{}l}{\emph{Choice}} \\
\quad Phase recognition & 907 & 1,167 &  & 921 &  &  & 1,443 & 1,361 &  &  &  &  &  &  & 1,104 &  &  &  & 4,686 & 691 & 1,526 \\
\quad Step recognition &  &  &  &  &  &  & 2,339 &  &  &  &  &  &  &  &  &  &  &  & 1,889 & 150 & 300 \\
\quad Action recognition &  &  &  &  &  &  &  &  &  & 1,259 &  &  &  &  &  &  & 1,933 &  & 2,363 & 265 & 564 \\
\quad Next action & 627 &  &  &  &  &  &  &  &  & 1,211 &  &  & 338 &  & 248 & 56 & 1,144 &  & 2,640 & 330 & 654 \\
\quad Next phase &  & 1,545 &  & 328 &  &  & 620 & 672 &  &  &  &  &  &  & 24 &  &  &  & 2,088 & 319 & 782 \\
\quad Next step &  &  &  &  &  &  & 1,569 &  &  &  &  &  &  &  &  &  &  &  & 1,119 & 150 & 300 \\
\quad Colon scene questions &  &  &  &  &  &  &  &  &  &  &  & 4,050 &  &  &  &  &  &  & 2,850 & 550 & 650 \\
\quad Polyp presence &  &  &  &  &  &  &  &  &  &  &  & 1,263 &  &  &  &  &  &  & 827 & 136 & 300 \\
\quad Histology &  &  &  &  &  &  &  &  &  &  &  & 97 &  &  &  &  &  &  & 57 & 8 & 32 \\
\midrule
\multicolumn{22}{@{}l}{\emph{Set}} \\
\quad Instrument set & 1,486 & 1,972 &  &  &  &  &  &  & 1,431 & 1,104 &  &  &  &  &  &  &  &  & 4,367 & 507 & 1,119 \\
\quad Action set & 1,500 &  &  &  &  &  &  &  & 1,431 &  &  &  &  &  &  &  &  &  & 2,059 & 272 & 600 \\
\quad Triplet set & 1,511 &  &  &  &  &  &  &  & 1,430 &  &  &  &  &  &  &  &  &  & 2,069 & 272 & 600 \\
\quad Segment set &  &  &  &  &  &  &  &  &  &  &  & 1,263 &  &  &  &  &  &  & 827 & 136 & 300 \\
\quad Scene graph &  &  &  &  &  &  &  &  &  &  &  &  &  &  &  &  &  & 1,338 & 969 & 150 & 219 \\
\quad Adverse-event detection &  &  &  &  &  &  & 670 &  &  &  &  &  &  &  &  &  &  &  & 404 & 96 & 170 \\
\quad Adverse-event severity &  &  &  &  &  &  & 335 &  &  &  &  &  &  &  &  &  &  &  & 202 & 48 & 85 \\
\quad Next camera motion &  &  &  &  &  &  &  & 235 &  &  &  &  &  &  &  &  &  & 244 & 333 & 86 & 60 \\
\midrule
\multicolumn{22}{@{}l}{\emph{Ordered}} \\
\quad Step sequence &  &  &  &  &  &  & 1,356 &  &  &  &  &  &  &  &  &  &  &  & 906 & 150 & 300 \\
\midrule
\multicolumn{22}{@{}l}{\emph{Component}} \\
\quad CVS criteria (clip) &  & 2,038 &  &  &  &  &  &  &  &  &  &  &  &  &  &  &  &  & 1,588 & 150 & 300 \\
\quad CVS criteria (frame) &  &  & 2,518 &  &  &  &  &  &  &  &  &  &  &  &  &  &  &  & 2,092 & 139 & 287 \\
\quad Skill rating &  &  &  &  &  &  &  &  &  &  & 206 &  &  &  &  &  &  &  & 156 & 18 & 32 \\
\midrule
\multicolumn{22}{@{}l}{\emph{Trajectory}} \\
\quad CVS criterion onset &  &  & 840 &  &  &  &  &  &  &  &  &  &  &  &  &  &  &  & 644 & 59 & 137 \\
\midrule
\multicolumn{22}{@{}l}{\emph{Timed caption}} \\
\quad Dense captioning & 1,774 & 936 &  & 244 &  &  & 1,280 & 272 & 1,007 & 925 &  & 329 & 1,183 &  & 1,493 & 2,247 & 3,408 & 457 & 13,231 & 823 & 1,501 \\
\quad Multi-view dense captioning &  &  &  &  &  &  &  &  &  &  &  &  &  &  &  &  &  & 253 & 208 & 11 & 34 \\
\midrule
\multicolumn{22}{@{}l}{\emph{Caption}} \\
\quad Video summary & 1,709 &  &  &  &  &  &  &  & 263 &  &  &  & 1,774 &  &  &  &  & 118 & 3,434 & 214 & 216 \\
\quad Polyp summary &  &  &  &  &  &  &  &  &  &  &  & 1,179 &  &  &  &  &  &  & 858 & 136 & 185 \\
\quad Region caption &  &  &  &  &  & 466 &  &  &  &  &  & 79 & 251 &  & 2,076 &  &  &  & 2,581 & 137 & 154 \\
\quad Multi-view summary &  &  &  &  &  &  &  &  &  &  &  &  &  &  &  &  &  & 117 & 106 & 4 & 7 \\
\midrule
\multicolumn{22}{@{}l}{\emph{Scalar (auxiliary)}} \\
\quad Polyp size &  &  &  &  &  &  &  &  &  &  &  & 78 &  &  &  &  &  &  & 47 & 4 & 27 \\
\midrule
\textbf{Total} & 11,538 & 9,331 & 5,675 & 2,278 & 2,211 & 1,446 & 11,749 & 3,994 & 8,541 & 6,704 & 206 & 9,430 & 4,663 & 253 & 7,177 & 3,560 & 9,092 & 4,228 & \textbf{77,261} & \textbf{8,454} & \textbf{16,361} \\
\bottomrule
\end{tabular*}
\end{table*}

\begin{table*}[t]
\centering
\caption{\textbf{Questions per task in the test-only suites}: blind windows by source dataset (left) and the four OOD suites (right).}
\label{tab:task_matrix_ood}
\scriptsize
\setlength{\tabcolsep}{2pt}
\begin{tabular*}{0.8\textwidth}{@{\extracolsep{\fill}}l rrrrrrrrrr rrrrr@{}}
\toprule
 & \multicolumn{10}{c}{Blind windows} & \multicolumn{5}{c}{OOD} \\
\cmidrule(lr){2-11}\cmidrule(l){12-16}
Task & \rotatebox{90}{AutoLaparo} & \rotatebox{90}{Cholec80-CVS} & \rotatebox{90}{CholecT50} & \rotatebox{90}{EgoSurgery} & \rotatebox{90}{m2cai16} & \rotatebox{90}{MultiBypass140} & \rotatebox{90}{ProstaTD} & \rotatebox{90}{REAL-colon} & \rotatebox{90}{SAR-RARP50} & \rotatebox{90}{\textbf{Total}} & \rotatebox{90}{HeiChole} & \rotatebox{90}{MB140 Strasbourg} & \rotatebox{90}{KUMC} & \rotatebox{90}{PhaKIR} & \rotatebox{90}{\textbf{Total}} \\
\midrule
\multicolumn{16}{@{}l}{\emph{Temporal}} \\
\quad Action localization & 300 & 300 & 300 & 300 & 300 & 300 & 265 &  & 50 & 2,115 &  &  &  &  &  \\
\quad Polyp visibility intervals &  &  &  &  &  &  &  & 153 &  & 153 &  &  &  &  &  \\
\midrule
\multicolumn{16}{@{}l}{\emph{Spatial}} \\
\quad Spatio-temporal grounding &  &  &  &  &  &  &  &  &  &  &  &  & 48 & 300 & 348 \\
\midrule
\multicolumn{16}{@{}l}{\emph{Choice}} \\
\quad Phase recognition & 300 & 300 & 300 & 279 & 300 & 294 &  &  &  & 1,773 & 691 &  &  & 300 & 991 \\
\quad Step recognition &  &  &  &  &  & 293 &  &  &  & 293 &  & 449 &  &  & 449 \\
\quad Next phase & 280 & 289 &  &  & 300 & 295 &  &  &  & 1,164 & 366 & 143 &  & 84 & 593 \\
\quad Next step &  &  &  &  &  & 297 &  &  &  & 297 &  & 449 &  &  & 449 \\
\midrule
\multicolumn{16}{@{}l}{\emph{Set}} \\
\quad Instrument set &  & 300 & 300 &  &  &  & 276 &  & 300 & 1,176 & 350 &  &  & 156 & 506 \\
\quad Action set &  &  & 300 &  &  &  & 193 &  &  & 493 & 300 &  &  &  & 300 \\
\quad Adverse-event detection &  &  &  &  &  &  &  &  &  &  &  & 124 &  &  & 124 \\
\quad Adverse-event severity &  &  &  &  &  &  &  &  &  &  &  & 62 &  &  & 62 \\
\midrule
\multicolumn{16}{@{}l}{\emph{Ordered}} \\
\quad Step sequence &  &  &  &  &  &  &  &  &  &  &  & 145 &  &  & 145 \\
\midrule
\multicolumn{16}{@{}l}{\emph{Timed caption}} \\
\quad Dense captioning & 87 & 184 & 86 &  & 76 & 241 &  &  & 154 & 828 & 100 & 143 &  & 50 & 293 \\
\midrule
\textbf{Total} & 967 & 1,373 & 1,286 & 579 & 976 & 1,720 & 734 & 153 & 504 & \textbf{8,292} & 1,807 & 1,515 & 48 & 890 & \textbf{4,260} \\
\bottomrule
\end{tabular*}
\end{table*}

\subsection{Construction}

\noindent\textbf{Prompt and label corrections.}
Questions do not reveal where a clip lies in its recording: all times are relative to the clip. Step and triplet questions list the full task vocabulary, and next-action questions whose labels were offset by one step are withdrawn. Other corrections concern next-phase questions whose frames already show the answer phase, colonoscopy visibility intervals that omitted non-polyp lesions, and placeholder segment labels. CoPESD segment labels and Cholec80 CVS intervals come from curated releases of the original annotations. Endoscapes experts rated 11,090 of its 58,813 frames, and frame-level CVS questions use only these rated frames.

\noindent\textbf{Splits.}
We define our own recording-disjoint splits, which differ from the official splits of CholecTrack20, Endoscapes, and EgoExOR. Empty answers make up 22--34\% of temporal training answers in every task and dataset (27\% overall). In five set-valued tasks dominated by a few answers, training and test questions are resampled so that no answer set exceeds 10\% of its task and dataset; a linear program chooses the quotas that minimize the F1 of the best constant answer.

\noindent\textbf{Sampling.}
Captions make up about a quarter of training questions. The remaining questions are balanced across tasks and datasets with temperature $\alpha = 0.5$ and at least 60 questions each where available, and each recording's contribution is capped so that long videos do not dominate. In validation and test, no recording supplies more than 40\% of the headline questions of a task and dataset. Questions that were not sampled remain available as extended splits. Besides the 15,101 headline questions, the test split has 610 questions from tasks with fewer than 60 questions or a single recording, and 650 colonoscopy scene questions from one recording. Clips have at most 180 frames. Dense captions have at most eight segments, each with clip-relative start and end times. Point events count as matched when the predicted midpoint falls within 0.5\,s.

\begin{figure*}[t]
\centering
\scriptsize
\newcommand{\flagpanel}[5]{%
\begin{minipage}[t]{0.19\linewidth}\raggedright
#1\\[2pt]
\textbf{#2} #3\\[1pt]
``#4''\\[2pt]
\textbf{Frames show:} #5
\end{minipage}}
\flagpanel{\includegraphics[width=\linewidth]{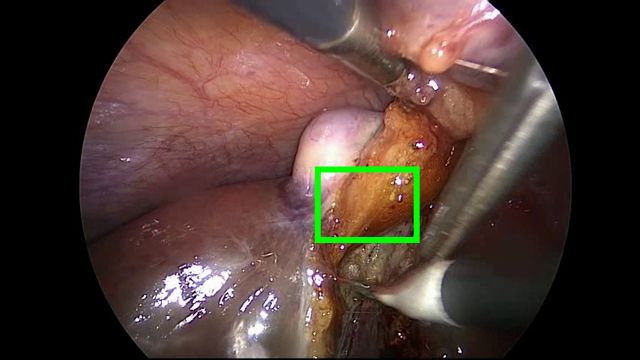}}{(a)}{Region caption}%
{\red{The hook}, positioned mostly in the upper right to center of the view, repeatedly dissects tissue from the gallbladder wall \ldots}%
{the green box contains only tissue; no instrument is inside it.}\hfill
\flagpanel{\includegraphics[width=\linewidth]{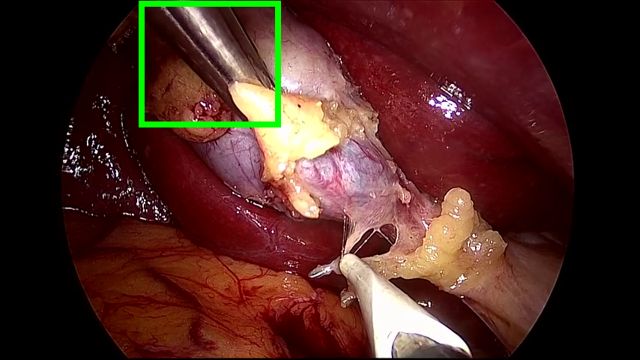}}{(b)}{Region caption}%
{\red{The hook}, positioned on the right side and center of the frames, steadily advances along the gallbladder wall \ldots}%
{the box is on the metal grasper at the upper left; the hook is the white-tipped instrument at the bottom.}\hfill
\flagpanel{\includegraphics[width=\linewidth]{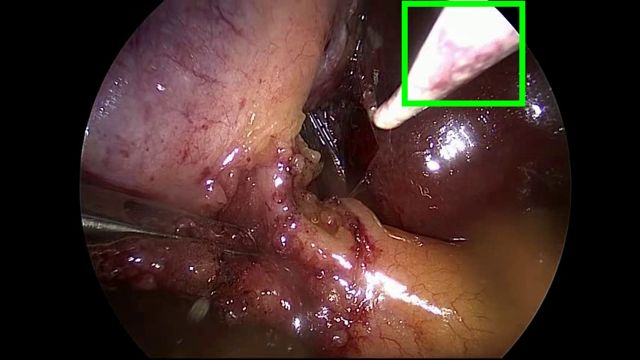}}{(c)}{Region caption}%
{Located on the left side of the field, \red{the grasper} consistently grasps and retracts tissue \ldots}%
{the box is on the white hook that enters from the right, not on the grasper at the left.}\hfill
\flagpanel{\includegraphics[width=\linewidth]{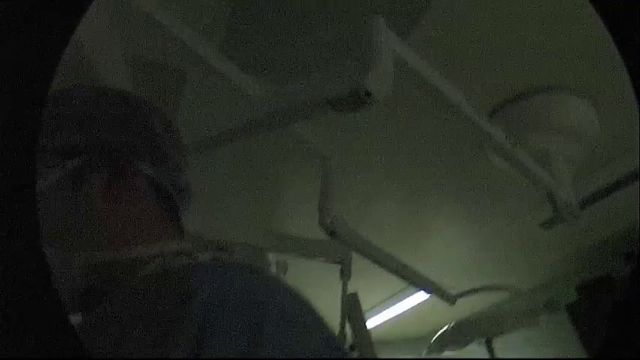}}{(d)}{Timed caption, 4:13--4:56}%
{Cleaning and coagulation with no annotated instrument visible, \red{the laparoscopic view remaining on the operative area}.}%
{at 4:54 the camera is already outside the body and shows the operating room.}\hfill
\flagpanel{\includegraphics[width=\linewidth]{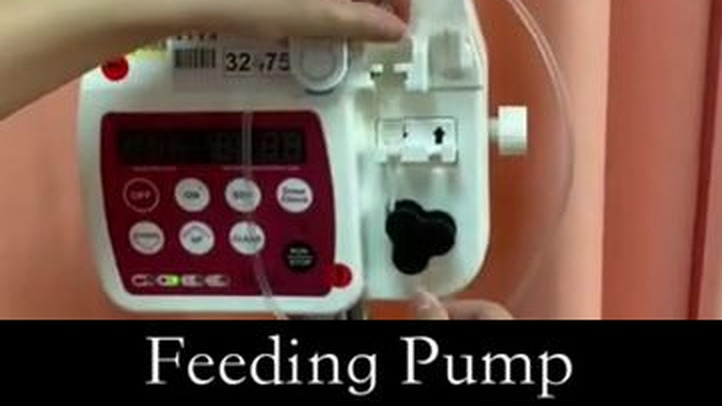}}{(e)}{Timed caption}%
{\ldots\ insert a segment of infusion tubing into the designated channel and roller mechanism of \red{an infusion pump} mounted on an IV pole \ldots}%
{the video teaches tube feeding into the gut with a feeding pump, not an intravenous infusion.}
\caption{\textbf{Captions flagged by the audit}, with the flagged claim in red. (a--c) Original MedVidBench region captions must describe the object in the green box on the first frame, and each names a different object. (d, e) \bench captions; the second reviewer rated both flags as minor.}
\label{fig:caption_flags}
\end{figure*}

\noindent\textbf{Caption audit.}
The original MedVidBench captions were written by GPT-4.1 and Gemini-2.5-Flash~\cite{su2026medgrpo}. The reviewer is a later model that wrote none of the captions (GPT-6.1-Sol, high reasoning); it checks each caption against up to 12 sampled frames per clip (\cref{tab:caption_audit}). A substantive error concerns an instrument, action, anatomy, boxed object, or segment content. A flag counts as an error only if a second reviewer, asked to defend the caption with the same frames, also finds it wrong; 86 of 102 first-pass flags were upheld. \Cref{fig:caption_flags} shows flagged captions from both sources.

\begin{table}[t]
\centering
\caption{\textbf{Independent caption audit.} Percent of judged items with a substantive error that a second reviewer upheld (number of judged items); rates are lower bounds.}
\label{tab:caption_audit}
\scriptsize
\setlength{\tabcolsep}{2.4pt}
\begin{tabular}{@{}l ccc c@{}}
\toprule
Captions & Summary & Region & Timed & All [95\% CI] \\
\midrule
MedVidBench, GPT & 1.7 (117) & 18.4 (49) & 21.1 (142) & 13.3 [10.0, 17.6] \\
MedVidBench, Gemini & 1.5 (133) & 14.5 (55) & 20.1 (144) & 11.7 [8.7, 15.7] \\
\bench test & 0.8 (128) & 0.0 (29) & 0.6 (165) & 0.6 [0.2, 2.2] \\
\bench train & 0.0 (48) & 4.2 (48) & 4.0 (50) & 2.7 [1.1, 6.8] \\
\bottomrule
\end{tabular}
\end{table}

\subsection{Video dependence per family}
\begin{figure}[t]
\centering
\includegraphics[width=\linewidth]{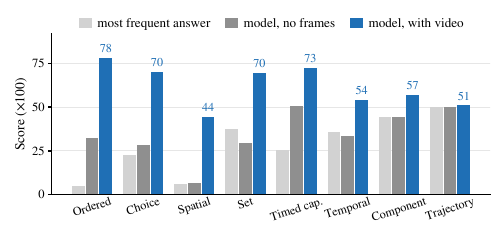}
\caption{\textbf{Video dependence per family.} The final 4B model with and without frames, and the most frequent training answer. Families are sorted by the margin that video adds.}
\label{fig:videofree}
\end{figure}

\Cref{fig:videofree} compares the final 4B model with and without frames against the most frequent training answer per family.

\subsection{Unseen-center suites}
\label{sec:supp_ood}
\Cref{tab:ood} reports every column of the four unseen-center suites; \cref{tab:main} shows five of them.
HeiChole anticipation clips have 32 frames and end one or five seconds before a phase transition. Its set clips are chosen so that no single answer dominates; without this selection, one constant instrument set would reach 78.8 F1. Its dense-caption windows allow phase changes anywhere in the clip.
Strasbourg step questions list all 46 steps without naming the phase.
PhaKIR grounding asks about the last of 16 frames; the suite uses the Cholec80 and CholecInstanceSeg question templates and restricts instruments to their vocabulary.

\begin{table*}[t]
\centering
\caption{\textbf{Out-of-distribution suites} ($\times100$), all absent from \bench training. T-cap: timed-caption F1; Gr.: grounding IoU. First row: best video-free answer; reference models: one-shot. Bold: block best; \colorbox{green!18}{shaded}: significantly above the SFT baseline; \colorbox{red!15}{shaded}: significantly below.}
\label{tab:ood}
\footnotesize
\setlength{\tabcolsep}{3.4pt}
\begin{tabular}{@{}l ccc cccc c cccc@{}}
\toprule
 & \multicolumn{3}{c}{HeiChole} & \multicolumn{4}{c}{Strasbourg} & KUMC & \multicolumn{4}{c}{PhaKIR} \\
\cmidrule(lr){2-4}\cmidrule(lr){5-8}\cmidrule(lr){9-9}\cmidrule(l){10-13}
Model & Choice & Set & T-cap & Choice & Set & Ord. & T-cap & Gr. & Choice & Set & Gr. & T-cap \\
\midrule
Video-free answer & 31.0 & 17.5 & 44.9 & 5.4 & 48.5 & 9.0 & 47.2 & 3.2 & 15.9 & 48.1 & 17.4 & -- \\
\midrule
InternVL3.5-8B & 22.7 & 40.0 & 7.0 & 8.0 & 27.6 & 3.1 & 5.0 & 14.2 & 26.6 & 55.2 & 13.6 & 14.9 \\
Hulu-Med-7B & 26.8 & 50.5 & 0.1 & 4.8 & 48.7 & 2.0 & 0.0 & 0.2 & 26.4 & 59.2 & 15.9 & 0.0 \\
MedGRPO-4B & 18.2 & 50.6 & 51.9 & 3.2 & 43.1 & 3.8 & 57.5 & 23.6 & 21.1 & 54.6 & 30.1 & 65.1 \\
Qwen3.5-4B & 15.6 & 53.6 & 5.4 & 6.2 & 39.7 & 2.4 & 6.0 & 9.5 & 25.5 & 30.4 & 8.8 & 28.7 \\
Qwen3.8-27B & 21.3 & 53.8 & 29.6 & 8.1 & 44.2 & 9.8 & 52.5 & 26.6 & 26.4 & 62.2 & 27.0 & 36.0 \\
Gemini 3.8 Flash & 26.0 & 41.5 & 41.2 & 4.0 & 41.4 & 6.1 & 37.8 & 17.9 & 11.3 & 44.4 & 13.9 & 46.6 \\
Gemini 4 Argon & 22.8 & 39.8 & 44.3 & 4.8 & 38.0 & 6.1 & 37.1 & 11.7 & 16.7 & 46.3 & 21.4 & 38.8 \\
GPT-6 Luna & 48.9 & 56.1 & 55.4 & 13.5 & 53.1 & 21.9 & 63.3 & 36.8 & 46.6 & 53.7 & 37.0 & 67.2 \\
\midrule
SFT baseline, 4B & 61.6 & \best{59.8} & 61.8 & 35.1 & 63.7 & 44.2 & \best{71.4} & 33.3 & 64.3 & \best{63.9} & 33.9 & 76.0 \\
\quad + QS, LQR & 62.4 & 59.2 & \best{65.2} & \gain 37.3 & 67.3 & \best{44.7} & 71.2 & 33.5 & 64.8 & 62.7 & \gain 72.2 & \best{77.6} \\
\quad\quad + RL & \best{63.5} & 59.1 & 63.3 & \gain \best{37.7} & \best{67.5} & 44.2 & 70.8 & \gain \best{38.9} & \best{65.2} & 59.1 & \gain \best{72.5} & \loss 73.1 \\
\midrule
SFT baseline, 27B & 63.0 & \best{58.0} & \best{67.5} & 34.4 & 65.7 & 47.9 & 70.1 & 42.3 & \best{65.9} & 59.0 & 42.0 & \best{80.7} \\
\quad + QS, LQR & 63.8 & 56.4 & \loss 63.1 & 35.1 & 70.6 & \gain \best{53.4} & 70.0 & \gain 49.7 & 62.8 & \best{63.6} & \gain 75.0 & 79.8 \\
\quad\quad + RL & \best{63.9} & 56.1 & 66.3 & \best{36.0} & \best{70.7} & 52.1 & \best{70.6} & \gain \best{50.8} & 63.9 & 61.1 & \gain \best{75.2} & 78.3 \\
\bottomrule
\end{tabular}
\end{table*}

\subsection{Blind-window test}
\label{sec:supp_realistic}
The blind-window test uses only test recordings and covers 33 combinations of task and dataset. For each, it keeps the window length, frame schedule, and prompt template of the curated test, and clips are cut by regular tiling instead of curation. For recognition, the reference is the label at the final frame of the window; for anticipation, that frame is the cutoff. Windows with more segments than the task allows are dropped. Each combination has at most 300 questions, sampled evenly across recordings without class balancing or selection of rare labels. Dense-caption references use the datasets' own templates, so segment timing is comparable with the curated test but caption wording is not. The 1,622 blind-window training questions are cut the same way from training recordings.

\noindent\textbf{Questions behind each score.}
Of the 8,292 questions, 2,268 are temporal: 515 windows contain the queried event and 1,753 do not. Presence is scored on all 2,268, and localization F1 on the 515 that contain the event. Boundary F1 is scored on the 286 of these with an event boundary inside the window, and dense-caption F1 on the 626 of 828 dense-caption windows that contain a phase change. The other questions are 2,066 on phase and step recognition, 1,461 on anticipation, and 1,669 on instrument and action sets.

\subsection{Question format and examples}
\label{sec:supp_examples}
Every question is the clip followed by a task specification, a JSON object with three parts: the question (instruction, constraints, and candidate labels), the answer format, and the clip state (duration and the timestamp of every frame). \Cref{fig:prompt_spec} shows a complete question, and \cref{tab:prompt_examples} gives one test question for each answer family.
For phase and step recognition, the reference is the native label at the final supplied frame.

\begin{figure}[t]
{\scriptsize
\begin{verbatim}
<video>
Task specification:
{"answer_format":{
   "syntax":"One option name, verbatim."},
 "questions":{"answer":{
   "constraints":["Select exactly one listed phase.
     The label applies to this supplied clip."],
   "criteria":{"options":["Preparation",
     "Dividing Ligament and Peritoneum",
     "Dividing Uterine Vessels and Ligament",
     "Transecting the Vagina","Specimen Removal",
     "Suturing","Washing"]},
   "instructions":"Which surgical phase is present
     in the observed clip?",
   "type":"choice"}},
 "state":{"camera_layout":"Single view.",
   "clip_duration_seconds":10.0,
   "frame_timestamps_seconds":[0.0,1.0,2.0,3.0,4.0,
     5.0,6.0,7.0,8.0,9.0],
   "input_span_seconds":[0.0,9.0],
   "media":"One chronological laparoscopic video
     frame sequence.",
   "time_coordinate_end_seconds":10.0,
   "time_origin":"First supplied frame is 0 seconds.
   All queries and answer times use this clip clock."}}
\end{verbatim}
}
\caption{\textbf{A complete question.} AutoLaparo phase recognition, 10 frames. Answer: ``Dividing Ligament and Peritoneum''.}
\label{fig:prompt_spec}
\end{figure}

\begin{table*}[t]
\centering
\caption{\textbf{One test question per answer family.} The question column gives the instruction, the main constraints, and the answer syntax from the task specification, with long option lists shortened; the reference column gives the scored answer.}
\label{tab:prompt_examples}
\scriptsize
\setlength{\tabcolsep}{4pt}
\begin{tabular}{@{}p{2.5cm} p{8.3cm} p{5.6cm}@{}}
\toprule
Family, task, dataset & Question & Reference answer \\
\midrule
Temporal: phase localization, Cholec80-CVS (64 frames) &
Locate every interval of the ClippingCutting surgical phase within the supplied clip. If the queried phase does not occur within this clip, return none. \emph{Syntax:} start-end seconds; multiple intervals separated by commas; empty answer: none. &
32-64 seconds \\
\addlinespace
Spatial: grounding, AutoLaparo (6 frames) &
Within the requested interval, where is the grasping forceps at the requested timestamps? Query times: 4.0, 5.0 seconds. \emph{Syntax:} t seconds: [x1, y1, x2, y2], one entry per timestamp; coordinates in 0--1000. &
4 seconds: [0, 406, 159, 628] \newline 5 seconds: [0, 426, 276, 650] \\
\addlinespace
Choice: phase recognition, AutoLaparo (10 frames) &
Which surgical phase is present in the observed clip? Options: Preparation; Dividing Ligament and Peritoneum; Dividing Uterine Vessels and Ligament; Transecting the Vagina; Specimen Removal; Suturing; Washing. \emph{Syntax:} one option name, verbatim. &
Dividing Ligament and Peritoneum \\
\addlinespace
Set: instrument set, Cholec80-CVS (32 frames) &
Which surgical instruments are visible at any supplied timestamp? Options: Grasper, Bipolar, Hook, Scissors, Clipper, Irrigator, SpecimenBag. \emph{Syntax:} comma-separated labels; empty set: none. &
Bipolar, Grasper, SpecimenBag \\
\addlinespace
Ordered: step sequence, MultiBypass140 (30 frames) &
List the distinct non-null surgical steps observed in the supplied clip, in order of first occurrence. Options: adhesiolysis, alimentary limb measurement, \ldots\ (46 step names). \emph{Syntax:} comma-separated step names; empty sequence: none. &
omental lifting, omental section \\
\addlinespace
Component: CVS at a frame, Endoscapes (16 frames) &
At the final timestamp, assess each Critical View of Safety component and whether CVS is achieved. C1: two and only two tubular structures, the cystic duct and cystic artery, enter the gallbladder. C2: the hepatocystic triangle is cleared of fat and fibrous tissue. C3: the lower third of the gallbladder is dissected off the cystic plate. \emph{Syntax:} C1: met/not met; C2: met/not met; C3: met/not met; CVS: achieved/not achieved. &
C1: not met; C2: not met; C3: met; CVS: not achieved \\
\addlinespace
Trajectory: CVS over a video, Endoscapes (16 frames, 80 s) &
When is each CVS criterion first observed as met at the supplied timestamps, and what is the final CVS status? (Criteria C1--C3 as above.) \emph{Syntax:} C1 first met \textless time\textgreater s/never; C2 \ldots; C3 \ldots; CVS achieved/never; final: achieved/not achieved. &
C1 first met 0s; C2 first met 25s; C3 first met 20s; CVS achieved; final: not achieved \\
\addlinespace
Timed caption: dense captioning, Cholec80-CVS (180 frames, 151 s) &
Describe the surgical phases and instruments in use in the supplied clip with their time spans. Return at most eight temporal caption segments. Caption each segment as ``\textless phase\textgreater\ with \textless instruments\textgreater'' using Cholec80 phase and tool names. \emph{Syntax:} start-end seconds: description, one segment per line. &
0-20 seconds: Gallbladder dissection with hook: the hook enters from the right and divides the thin connective tissue between the gallbladder and the organ surface below it. \newline 20-31 seconds: Gallbladder dissection with grasper and hook: a grasper at the upper right holds the gallbladder while the hook separates the thin tissue plane beneath it. \newline \ldots\ (further segments) \\
\addlinespace
Caption: video summary, CholecT50 (96 frames, 268 s) &
Summarize the video content in a short paragraph, focusing on the main surgical activities. \emph{Syntax:} plain text. &
Graspers first place the gallbladder into a specimen bag and close it. A grasper then retracts the liver, followed by extended retraction of the omentum and aspiration of fluid with an irrigator. Finally, graspers pull the specimen bag holding the gallbladder by its closure string, drawing it upward and out of the abdominal cavity until the endoscope is withdrawn. \\
\bottomrule
\end{tabular}
\end{table*}

\section{Training and evaluation details}
\label{sec:supp_rl}

\noindent\textbf{Supervised training.}
All supervised epochs use a global batch of 64; the vision encoder and projector stay frozen. The first two epochs use learning rate $10^{-5}$. The third epoch starts from the selected two-epoch checkpoint and runs 1,208 optimizer steps at learning rate $2.5\times10^{-6}$. The SFT baseline follows the same three-epoch schedule and differs only in prompts and inputs, so the reported gains do not include the effect of further training. Query-slot inputs apply to grounding, frame-level CVS, and scene-graph questions in training and inference; the rows ``QS at inference'' and ``QS, 1$\times$ px'' in \cref{tab:ablation} use frame replacement on grounding questions only. Frames have at most 147,456 pixels by default and 294,912 in spatial tasks (grounding, scene graphs, region captions, and multi-view questions); query-slot grounding frames use up to 589,824 pixels, twice the spatial budget. All frames of a clip share a budget of 8,388,608 pixels, which limits long clips, so grounding frames are 1.35$\times$ larger on average.

\noindent\textbf{Reinforcement learning.}
DAPO~\cite{yu2025dapo} samples 8 answers per question at temperature 0.8 and top-$p$ 0.95, with at most 8,192 prompt tokens and 768 response tokens. It uses group-relative advantages, asymmetric clipping ($\epsilon_{\text{low}}=0.2$, $\epsilon_{\text{high}}=0.28$), no KL penalty, AdamW with weight decay 0.01 and learning rate $10^{-6}$, and gradient-norm clipping at 0.5.
The temporal reward averages interval F1 over tIoU thresholds $\{0.3, 0.5, 0.7\}$. The caption judge (Qwen3.8-Flash-Next) scores correctness, completeness, faithfulness, terminology, and task focus against the reference.
For multiple-choice tasks, each answer's normalized advantage is multiplied by $w = 1/(K p_c)$, where $K$ is the number of classes and $p_c$ the training frequency of the reference class within its task and dataset; weights are capped at 3 and renormalized to mean one. The weight is applied to advantages because a weight on rewards would cancel under group normalization. Group filtering uses unweighted correctness. CVS and skill questions, whose answers have several fields, use the mean class weight of their fields. For timed captions, the caption half of the reward is divided by the larger of the predicted and reference segment counts, so extra segments earn nothing.
The MedGRPO comparison applies the MedGRPO recipe~\cite{su2026medgrpo} to the SFT baseline: rewards rescaled by percentiles of the supervised model, a format factor, GRPO without group filtering, and the original questions, for 300 steps at learning rate $5\times10^{-7}$.

\noindent\textbf{Question pools.}
The \method pool has the same 77,261 questions in total and the same number for every task and dataset. It replaces 31,383 of them (41\%): 30,163 unsampled questions and 1,220 re-windowed MultiBypass140 and Cholec80 phase-recognition clips.
Labels are matched by combination of criteria in the CVS tasks and by whether the event occurs in temporal tasks. Each class keeps its count and is filled with unseen questions first.
The pool for the supervised epoch also matches the number of answers per set (empty, one, \ldots, five or more), which leaves 30,733 unseen questions (40\%).

\noindent\textbf{Evaluation.}
\label{sec:supp_eval}
All models decode greedily with up to 1,024 new tokens. Caption scores come from the same judge as the caption reward, with reasoning off.
GPT-6 Luna (high reasoning, low image detail), Gemini 3.8 Flash and Gemini 4 Argon (both medium thinking), and InternVL3.5-8B receive the task prompt, one text-only training example, and up to 50 timestamped frames. Qwen3.5-4B, Qwen3.8-27B, and Hulu-Med-7B receive the same prompt and example with their native video input, which holds every frame of the clip (at most 180).
For the models limited to 50 frames, longer clips are subsampled evenly, and each frame is resized to at most 147,456 pixels, the default per-frame budget of our Qwen models. GPT-6 Luna stays below both trained models when every model receives every frame: on the 1,181 curated grounding questions with at most 50 frames, GPT-6 Luna scores 27.5, the SFT baseline 37.0, and the final 4B model 55.8. The released MedGRPO model, a Qwen3-VL-4B trained on MedVidBench, receives MedVidBench prompt templates on the tasks it shares with MedVidBench and the one-shot prompt on new tasks, the better protocol for each task group (\cref{sec:supp_medgrpo_prompts}).
Video-free baselines are constant answers, the most frequent training answer, equal clip segments, and the final 4B model without frames.
On blind windows, localization is scored on windows that contain the event, boundary F1 on windows with an event boundary inside, and dense-caption F1 on windows with a phase change.

\section{Additional results}
\subsection{Component analysis}
\label{sec:supp_ablation}

\noindent\textbf{Query-slot crossover.}
\label{sec:supp_qs_crossover}
\Cref{tab:ablation} shows that slots applied only at inference already help. The opposite case shows what slot training teaches: a model trained with slots but evaluated on standard inputs scores 31.3 on curated grounding and 36.2 on PhaKIR (4B, standard pixel budget), against 30.4 and 33.9 for the baseline and 36.8 and 62.9 for the same model with slots at inference. It keeps only 0.9 points of its grounding gain ([0.4, 1.3]) and 2.4 PhaKIR points, so slot training teaches the model to use the marked frame.

\noindent\textbf{Queried frames without the clip.}
\label{sec:supp_frame_only}
\Cref{tab:frame_only} uses the 1,514 of 1,584 curated grounding questions whose queried times are sampled frames, and all KUMC and PhaKIR questions. The other 70 are REAL-colon questions that ask about times between sampled frames. The clip is replaced with only the queried frames, at matched frame size; the question stays the same.

\begin{table}[t]
\centering
\caption{\textbf{\method across both stages} (4B, $\times100$). From the query-slot model: RL on the original questions, or \method in the third epoch and the RL pool, then RL. Str., PK: Strasbourg, PhaKIR. Bold: best.}
\label{tab:lqr_rl}
\scriptsize
\setlength{\tabcolsep}{1.0pt}
\begin{tabular*}{\linewidth}{@{\extracolsep{\fill}}l ccccc cc cc cc@{}}
\toprule
 & \multicolumn{5}{c}{Curated} & \multicolumn{2}{c}{Blind} & \multicolumn{2}{c}{Str.} & \multicolumn{2}{c}{PK} \\
\cmidrule(lr){2-6}\cmidrule(lr){7-8}\cmidrule(lr){9-10}\cmidrule(l){11-12}
 & Spat. & Temp. & Ch. & Set & T-cap & Pres. & Loc. & Ch. & Set & Spat. & Set \\
\midrule
QS & 41.2 & 52.9 & \best{70.7} & 69.5 & 69.9 & 81.2 & 50.3 & 35.0 & 64.2 & 70.2 & 64.3 \\
\quad + RL & 43.0 & 52.7 & 70.4 & 69.0 & 72.1 & \best{86.9} & 56.8 & 35.1 & 62.6 & 71.3 & \best{64.4} \\
\quad + LQR, RL (final) & \best{44.4} & \best{54.0} & 70.1 & \best{69.7} & \best{72.7} & 86.7 & \best{57.7} & \best{37.7} & \best{67.5} & \best{72.5} & 59.1 \\
\bottomrule
\end{tabular*}
\end{table}

\noindent\textbf{\method across both stages.}
\label{sec:supp_lqr_rl}
\Cref{tab:lqr_rl} separates the two parts of the final recipe after query slots. RL from the query-slot model with the original questions already brings the untrimmed-video gain: blind-window presence rises from 81.2 to 86.9 and localization F1 from 50.3 to 56.8. Adding \method, in the third epoch and in the RL pool, then improves evidence-dependent answers and transfer to Strasbourg: curated grounding rises by a further 1.4 points ([0.8, 2.0]), Strasbourg choice accuracy by 2.5 ([0.5, 4.8]), and Strasbourg set F1 by 4.8 ([0.3, 9.7]).

\noindent\textbf{Run-to-run variation.}
\label{sec:supp_noise}
With batched half-precision decoding, between two runs of the same checkpoint, family scores differ by 0.1 to 0.9 points, and by 1.5 on the trajectory family, which has 137 questions. We therefore treat differences below about one point as ties.

\subsection{MedVidBench original test}
\Cref{tab:medvidbench_test} evaluates all models on the original MedVidBench test questions with their original labels; none of the 142 test recordings appears in \bench training. MedVidBench SFT is the same Qwen3.5-4B backbone trained on MedVidBench. Every model answers the original MedVidBench prompts, which differ from \bench in wording and in clip length for frame-level tasks (128 frames on average for grounding, against 28 in \bench). The final 4B model has the best grounding IoU (31.8), and the 27B models lead temporal localization, dense-caption timing, CVS, and skill assessment. The 871 video-summary and region-caption questions need MedVidBench's own judge and are omitted, leaving 5,374 of 6,245.

\begin{table}[t]
\centering
\caption{\textbf{MedVidBench original test} (5,374 scored questions, $\times100$). TAL: temporal mIoU at tIoU 0.3 / 0.5; STG: box mIoU; NA: next action; DVC: dense-caption F1; CVS, Skill: accuracy. Bold: best.}
\label{tab:medvidbench_test}
\scriptsize
\setlength{\tabcolsep}{0pt}
\begin{tabular*}{\columnwidth}{@{\extracolsep{\fill}}l ccccccc@{}}
\toprule
Model & TAL@.3 & TAL@.5 & STG & NA & DVC & CVS & Skill \\
\midrule
MedVidBench SFT, 4B & 52.5 & 48.9 & 24.4 & \best{57.2} & 52.1 & 48.2 & 30.9 \\
MedGRPO-4B & 52.2 & 46.8 & 22.9 & 34.6 & 50.2 & 41.1 & 29.0 \\
\midrule
SFT baseline, 4B & 48.9 & 43.6 & 28.3 & 46.9 & 45.8 & 47.6 & 36.9 \\
\quad + QS, LQR & 50.9 & 45.5 & 29.6 & 50.4 & 46.3 & 47.9 & 34.9 \\
\quad\quad + RL & 52.8 & 47.3 & \best{31.8} & 47.3 & 48.9 & 48.1 & 36.2 \\
SFT baseline, 27B & 51.2 & 46.7 & 20.0 & 52.8 & 50.4 & 47.9 & 39.8 \\
\quad + QS, LQR & 53.6 & 49.1 & 21.5 & 56.0 & 52.1 & 48.7 & 39.1 \\
\quad\quad + RL & \best{54.7} & \best{49.9} & 22.9 & 54.0 & \best{53.9} & \best{50.9} & \best{40.4} \\
\bottomrule
\end{tabular*}
\end{table}

\renewcommand{\dblfloatpagefraction}{0.6}%
\begin{figure*}[p]
\centering
\includegraphics[width=\linewidth]{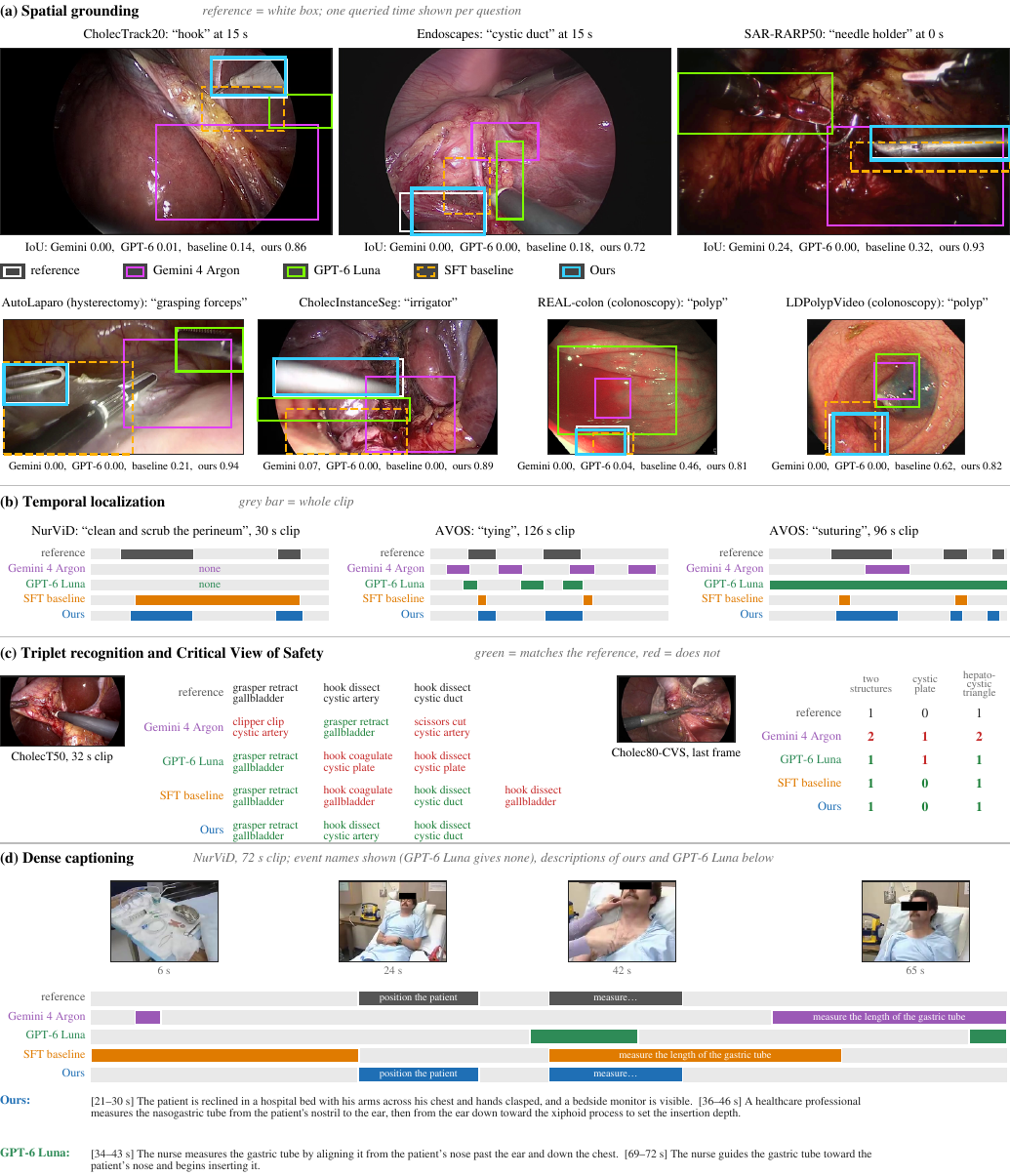}
\caption{\textbf{Qualitative comparison} with the SFT baseline, GPT-6 Luna, and Gemini 4 Argon on test questions. (a) Box IoU at one queried time. (b, d) Reference and predicted intervals; (d) also lists the event descriptions. (c) Items matching the reference in green, others in red.}
\label{fig:qual_suppl}
\end{figure*}

\begin{figure*}[t]
\centering
\includegraphics[width=\linewidth]{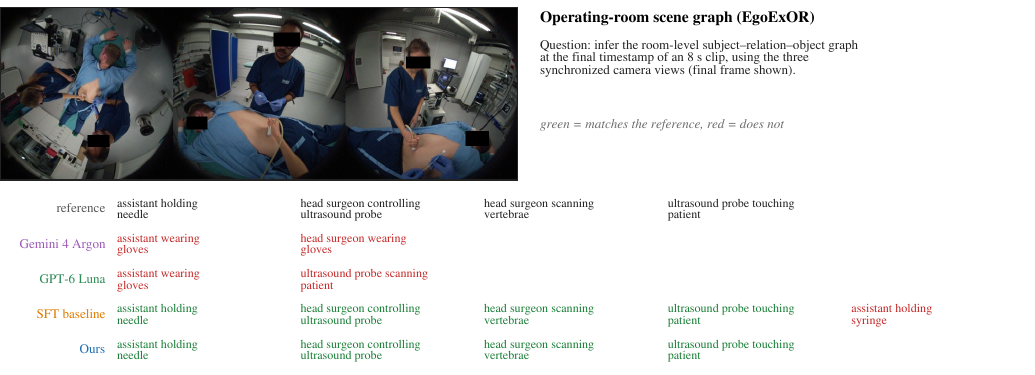}
\caption{\textbf{Operating-room scene graph.} Our model returns the four reference relations; the SFT baseline adds a relation that is not in the reference graph, and the commercial models return generic relations.}
\label{fig:sgg_suppl}
\end{figure*}

\subsection{Prompt protocol for the MedGRPO model}
\label{sec:supp_medgrpo_prompts}
We evaluated the released MedGRPO model under four prompt protocols: the \bench structured prompt without an example, the same prompt with one example, MedVidBench templates on all tasks, and MedVidBench templates on the tasks shared with MedVidBench combined with the one-shot prompt on new tasks. MedVidBench templates raise its scores on the shared tasks (timed-caption F1 of 50.7, against 40.9 with the one-shot prompt). On tasks new in \bench, the one-shot prompt scores far higher (trajectory score of 29.7, against 2.3 with MedVidBench templates). The main tables use the combination, which takes the better protocol for each task group. The SFT baseline exceeds the released model on every curated and unseen-hospital column of \cref{tab:main} under every protocol.

\subsection{Qualitative comparison}
\label{sec:supp_qual}
\Cref{fig:qual_suppl} compares our final 4B model with the SFT baseline, GPT-6 Luna, and Gemini 4 Argon on five task types. Across cholecystectomy, prostatectomy, hysterectomy, and colonoscopy, our model's boxes stay on the queried instrument, structure, or polyp, where the baseline's boxes drift and the commercial models box a neighbouring region. It recovers each occurrence of a repeated action, where the other models merge occurrences, miss them, or mark the whole clip. It returns the reference triplets, and both trained models match the Critical View of Safety reference, whereas the commercial models list other targets and score components that are not yet achieved. In dense captioning it finds both annotated events with their boundaries. \Cref{fig:sgg_suppl} adds an operating-room scene graph, where it returns exactly the annotated relations between staff, instruments, and patient.

\end{document}